\documentclass[11pt]{article}

\usepackage[a4paper,margin=25mm]{geometry}
\usepackage[T1]{fontenc}
\usepackage[utf8]{inputenc}
\usepackage{amsmath,amssymb}
\usepackage{newtxtext,newtxmath}
\usepackage{graphicx}
\usepackage{booktabs}
\usepackage{multirow}
\usepackage{url}
\usepackage[round,authoryear]{natbib}
\usepackage[labelfont=bf,font=small]{caption}
\usepackage[colorlinks=true,linkcolor=blue!50!black,citecolor=blue!50!black,urlcolor=blue!50!black]{hyperref}
\usepackage{float}

\graphicspath{{./}}
\newcommand{\PE}{\mathrm{PE}}
\newcommand{\Gtwo}{G2G}
\newcommand{\vtwo}{SoccerMap-v2}

\expandafter\def\csname supp@app:calibration\endcsname{S10}
\expandafter\def\csname supp@app:firstcycle\endcsname{S8}
\expandafter\def\csname supp@app:ladder\endcsname{S11}
\expandafter\def\csname supp@app:length\endcsname{S12}
\expandafter\def\csname supp@app:repro\endcsname{S13}
\expandafter\def\csname supp@app:signtests\endcsname{S9}
\expandafter\def\csname supp@fig:convergence\endcsname{S2}
\expandafter\def\csname supp@fig:paired\endcsname{S1}
\expandafter\def\csname supp@sec:convergence\endcsname{S5.4}
\expandafter\def\csname supp@sec:lrsweep\endcsname{S5.5}
\expandafter\def\csname supp@sec:multiseed\endcsname{S5.2}
\expandafter\def\csname supp@sec:paireddiff\endcsname{S5.3}
\expandafter\def\csname supp@sec:pooled\endcsname{S5.1}
\expandafter\def\csname supp@supp:compute\endcsname{S7}
\expandafter\def\csname supp@supp:designs\endcsname{S1}
\expandafter\def\csname supp@supp:external\endcsname{S2}
\expandafter\def\csname supp@supp:families\endcsname{S4}
\expandafter\def\csname supp@supp:gates\endcsname{S3}
\expandafter\def\csname supp@supp:halfrule\endcsname{S4.6.1}
\expandafter\def\csname supp@supp:history\endcsname{S5}
\expandafter\def\csname supp@supp:limitations\endcsname{S5.6}
\expandafter\def\csname supp@supp:tail\endcsname{S6}
\expandafter\def\csname supp@tab:ablation\endcsname{S19}
\expandafter\def\csname supp@tab:compute\endcsname{S35}
\expandafter\def\csname supp@tab:compute-audit\endcsname{S36}
\expandafter\def\csname supp@tab:cv\endcsname{S4}
\expandafter\def\csname supp@tab:cv-folds\endcsname{S5}
\expandafter\def\csname supp@tab:cv-raw\endcsname{S6}
\expandafter\def\csname supp@tab:cv-raw-folds\endcsname{S7}
\expandafter\def\csname supp@tab:ece\endcsname{S39}
\expandafter\def\csname supp@tab:epochs\endcsname{S46}
\expandafter\def\csname supp@tab:external-a\endcsname{S8}
\expandafter\def\csname supp@tab:external-b\endcsname{S9}
\expandafter\def\csname supp@tab:external-data\endcsname{S1}
\expandafter\def\csname supp@tab:external-fold1\endcsname{S11}
\expandafter\def\csname supp@tab:external-folds\endcsname{S10}
\expandafter\def\csname supp@tab:external-raw-a\endcsname{S12}
\expandafter\def\csname supp@tab:external-raw-b\endcsname{S13}
\expandafter\def\csname supp@tab:external-raw-fold1\endcsname{S15}
\expandafter\def\csname supp@tab:external-raw-folds\endcsname{S14}
\expandafter\def\csname supp@tab:firstcycle\endcsname{S37}
\expandafter\def\csname supp@tab:gate_audit\endcsname{S2}
\expandafter\def\csname supp@tab:gate_audit_bytype\endcsname{S3}
\expandafter\def\csname supp@tab:grad\endcsname{S43}
\expandafter\def\csname supp@tab:halfgrid\endcsname{S21}
\expandafter\def\csname supp@tab:hier-attrib\endcsname{S40}
\expandafter\def\csname supp@tab:hier-conv\endcsname{S41}
\expandafter\def\csname supp@tab:hyper\endcsname{S45}
\expandafter\def\csname supp@tab:inputs\endcsname{S44}
\expandafter\def\csname supp@tab:knockouts\endcsname{S25}
\expandafter\def\csname supp@tab:lr\endcsname{S32}
\expandafter\def\csname supp@tab:multiseed\endcsname{S29}
\expandafter\def\csname supp@tab:multiseed-perseed\endcsname{S30}
\expandafter\def\csname supp@tab:noise\endcsname{S26}
\expandafter\def\csname supp@tab:noise_ar1\endcsname{S27}
\expandafter\def\csname supp@tab:o1attrib\endcsname{S42}
\expandafter\def\csname supp@tab:paired-full\endcsname{S31}
\expandafter\def\csname supp@tab:perseed\endcsname{S18}
\expandafter\def\csname supp@tab:pooled\endcsname{S28}
\expandafter\def\csname supp@tab:raster\endcsname{S20}
\expandafter\def\csname supp@tab:setattn\endcsname{S23}
\expandafter\def\csname supp@tab:sign\endcsname{S38}
\expandafter\def\csname supp@tab:tail\endcsname{S34}
\expandafter\def\csname supp@tab:tuned\endcsname{S16}
\expandafter\def\csname supp@tab:tuned-raw\endcsname{S17}
\expandafter\def\csname supp@tab:tuned-setattn\endcsname{S24}
\expandafter\def\csname supp@tab:unet\endcsname{S22}
\expandafter\def\csname supp@tab:value\endcsname{S33}

\newcommand{\suppref}[1]{\ifcsname supp@#1\endcsname\csname supp@#1\endcsname\else\textbf{S??}\GenericWarning{}{LaTeX Warning: supplement reference `#1' undefined}\fi}

\title{Graph-to-Grid (G2G): Continuous-Coordinate Feature Painting\\
for Soccer Pass Surfaces}

\author{
Kaan G\"unay \\
\small Sabanc{\i} University, Istanbul, Turkey \\
\small \texttt{kaan@pittsburghdynamics.net} \\
\small \url{https://orcid.org/0009-0008-5517-1321}
\and
Orhun Gun\thanks{Corresponding author. \texttt{orhun@cmu.edu}} \\
\small Carnegie Mellon University, Pittsburgh, PA, USA \\
\small \texttt{orhun@cmu.edu} \\
\small \url{https://orcid.org/0009-0003-1578-7117}
}
\date{}

\begin{document}
\maketitle

\begin{abstract}
Dense pass surfaces give, for every pitch cell, whether a pass played there
would arrive, whether the carrier would choose it, and what the possession
would then be worth. The networks that draw them read the state as a raster
of per-cell counts, losing where inside a cell each player stands. LiDAR
detectors, bird's-eye-view perception and graph weather models move entity features onto a grid, binning each entity to a cell or learning the
transfer. We evaluate the interpolated form: each player's features are
scattered bilinearly onto the grid at the player's measured coordinates, so
the surface loss trains the per-player encoder end to end. Those systems
adopt an interface; this paper measures one. On
53{,}628 passes from the 2022 World Cup, painting improves selection
likelihood over the same core fed rasters alone by about a quarter of a nat:
in every match of an eight-fold cross-validation, with every arm tuned over
five seeds, and after retraining on seven Bundesliga and 2.~Bundesliga matches from
another provider. Thirteen pre-specified studies locate the gain: painting the nine
raw player features with no encoder carries three quarters of it, and the
learned encoder and message passing add a smaller, resolved increment.
Painting also helps the original SoccerMap and a canonical U-Net, whereas
offset channels, a finer raster, an attention painter and a raster-free
decoder do not. Frozen across the provider boundary the likelihood advantage
is lost; injected tracking error compresses it. These results concern
observed-endpoint prediction, not calibrated evaluation of hypothetical
passes.
\end{abstract}


\section{Introduction}
\label{sec:intro}

An analyst who pauses a match wants, for every spot on the pitch, three
numbers: if the ball were played there, would it arrive; would this carrier
choose it; and what would the possession be worth afterwards. Answering them
for every cell produces a surface over the pitch, and surfaces are what
practitioners read: they show the options a player had, not just the one
taken. Learning such surfaces from historical passes is hard because each
example supplies one observed endpoint, and an unsuccessful pass may end at
an interception rather than at its intended destination. In this paper,
``selection'' denotes prediction of the observed endpoint. Completion and
value are supervised at that endpoint; their values at unchosen locations
are model extrapolations, not validated counterfactuals.

SoccerMap-style networks represent the match state as a raster and predict a
dense pitch surface~\citep{fernandez2020soccermap,fernandez2021epv}. Occupancy
and aggregated velocity channels lose within-cell position and the
association between a player and that player's own features. Graph models
retain players and their relations, but the soccer systems reviewed here
predict over players or
events~\citep{stockl2021making,wang2024tacticai,rahimian2026tgn}, not over
pitch cells. A standard entity-to-grid interface connects the two: scatter
each entity's features bilinearly onto the grid at its measured continuous
coordinates, then process the grid with a convolutional network. Interfaces of this kind carry LiDAR object
detectors~\citep{lang2019pointpillars}, bird's-eye-view perception from
cameras~\citep{philion2020lss} and graph weather
models~\citep{lam2023graphcast}. They differ in where an entity's features
land: those three bin each entity to one cell or learn the transfer, whereas
the interface studied here writes at the measured continuous position, and
Sec.~\ref{sec:ablation} prices that difference.

We study this interface through the Graph-to-Grid (\Gtwo{}) family, our name
for a nested set of models whose arms differ only in what is painted onto a
shared raster-only core (Sec.~\ref{sec:method}, Fig.~\ref{fig:arch}; on the
name, Sec.~\ref{sec:related}). Bilinear scatter-add is established machinery, and its differentiability is
inherited. What has not been established is what the interface buys and
which part of it pays, and soccer pass surfaces are an unusually clean place
to find out: one observed endpoint supervises every example, and the same
core, corpus, recipe and evaluation code can be held fixed while only the
front end changes. The contribution is that measurement.

\subsection{Claims, evidence and importance}

The main empirical claim is that continuous-coordinate feature painting
improves observed-endpoint prediction over the tested raster-only systems.
The full front end improves selection negative log-likelihood (NLL) over
the same core by $0.251$ nats in eight-fold cross-validation over all 64
World Cup matches, $0.292$ nats under per-arm learning-rate selection over
five seeds, and $0.239$ nats after retraining on seven external German league matches, ahead in every match of each comparison (Table~\ref{tab:glance}).
These are complementary checks, not three independent replications:
cross-validation and tuning retain design choices made on the original
World Cup split, while the external corpus is from a different provider,
league and season.

The attribution is more specific than an improvement from graph learning.
Painting the nine raw node features with no learned encoder retains
approximately three quarters of the selection gain, on the World Cup and on
the external corpus alike; the learned encoder and message passing together
add a further $0.065$--$0.067$ nats in the cross-validation and tuned
comparisons. That increment is resolved there, over 64 match clusters and
five seeds, whereas the three-seed holdout studies alone could not
characterise it. Painting also improves the original SoccerMap and a
canonical U-Net, so the effect is not tied to one core. Sub-cell offset
channels, a raster four times finer, a learned attention painter and a
raster-free set decoder do not close the gap, within the scope stated in
Sec.~\ref{sec:threats}.

For a new deployment we would start from raw painting: it has no front-end
parameters, keeps most of the gain, and is the cheaper of the two painted
variants (Sec.~\ref{sec:compute}). We would add the learned front end where
in-distribution selection likelihood is the objective and the tracking
provider is fixed. Frozen across a provider boundary, the raster-only core
has the best selection likelihood of the three; raw painting is numerically
ahead of \Gtwo{} there, but the contrast does not cross the study's adjusted
max-$T$ threshold. Entity-to-grid interfaces are already load-bearing in the detection,
perception and forecasting systems cited above, which bin to a cell or learn
the transfer rather than writing at continuous coordinates. The attribution
reported here, that placement and the write carry most of the gain and the
learned content the rest, is the kind of question those systems leave open.
The present experiments establish evidence only for soccer endpoint
prediction.

\begin{table}[htb]
\caption{The interface effect at a glance. $\Delta$ is the selection NLL of
the first arm minus the second in nats, so a negative value favours the
first arm; 95\% intervals from each study's match-cluster or hierarchical
bootstrap. Arm~C is the raster-only core \vtwo{}; D adds the painted front
end with a per-node encoder; $\mathrm{D}_{\mathrm{raw}}$ paints the nine raw
node features with no encoder; \Gtwo{} replaces the per-node encoder with
the relational one; \emph{original} is the published SoccerMap;
U$_{\mathrm{raster}}$ and U$_{\mathrm{raw}}$ read the same two inputs with a
canonical U-Net core. Study numbers follow Table~\ref{tab:studymap}; every
full family is in the supplement}
\label{tab:glance}
\centering\scriptsize
\setlength{\tabcolsep}{3pt}
\begin{tabular}{llrl}
\toprule
Contrast & Study, Sec. & $\Delta$ selection NLL & Reading \\
\midrule
\multicolumn{4}{l}{\emph{Value of the painted write}} \\
\Gtwo{}\,$-$\,C, eight-fold CV over 64 matches & 5, \S\ref{sec:cv} & $-0.251$ $[-0.263,-0.239]$ & ahead in 64 of 64 matches \\
\Gtwo{}\,$-$\,C, every arm tuned, converged, 5 seeds & 4, \S\ref{sec:tuned} & $-0.292$ $[-0.338,-0.245]$ & margin survives per-arm tuning \\
\Gtwo{}\,$-$\,C, other provider, retrained & 7, \S\ref{sec:external} & $-0.239$ $[-0.285,-0.197]$ & ahead in 7 of 7 matches \\
original$_{\mathrm{splat}}$\,$-$\,original & 8, \S\ref{sec:raster} & $-0.295$ $[-0.416,-0.214]$ & same write on the published core \\
U$_{\mathrm{raw}}$\,$-$\,U$_{\mathrm{raster}}$ (canonical U-Net core) & 13, \S\ref{sec:raster} & $-0.242$ $[-0.264,-0.213]$ & same write on a third core \\
U$_{\mathrm{raster}}$\,$-$\,C (U-Net core, raster only) & 13, \S\ref{sec:raster} & $+0.366$ $[+0.333,+0.397]$ & our core reads better (shared recipe) \\
\midrule
\multicolumn{4}{l}{\emph{Attribution within the front end}} \\
D$_{\mathrm{raw}}$\,$-$\,C, eight-fold CV over 64 matches & 12, \S\ref{sec:cv} & $-0.186$ $[-0.199,-0.172]$ & raw painter carries three quarters \\
\Gtwo{}\,$-$\,D$_{\mathrm{raw}}$, eight-fold CV over 64 matches & 12, \S\ref{sec:cv} & $-0.065$ $[-0.078,-0.052]$ & learned content: the remaining quarter \\
D$_{\mathrm{raw}}$\,$-$\,C, tuned, converged, 5 seeds & 12, \S\ref{sec:tuned} & $-0.225$ $[-0.256,-0.195]$ & same share under tuning \\
D$_{\mathrm{raw}}$\,$-$\,C, other provider, retrained & 10, \S\ref{sec:external} & $-0.178$ $[-0.207,-0.139]$ & three quarters of the margin, no encoder \\
D$_{\mathrm{raw}}$\,$-$\,D (raw features for the embedding) & 3, \S\ref{sec:ablation} & $+0.005$ $[-0.049,+0.064]$ & no separation over 8 matches \\
\Gtwo{}\,$-$\,D (message passing on top of painting) & 1, \S\ref{sec:attribution} & $-0.0515$ $[-0.095,-0.005]$ & clears max-$T$; seed sd as large \\
D$_{\mathrm{nearest}}$\,$-$\,D (nearest cell for bilinear) & 3, \S\ref{sec:ablation} & $+0.084$ $[+0.033,+0.125]$ & placement and interpolation: half the package \\
\midrule
\multicolumn{4}{l}{\emph{Raster and learned-interface substitutes}} \\
C$_{\mathrm{offsets}}$\,$-$\,C (sub-cell offset channels) & 8, \S\ref{sec:raster} & $-0.044$ $[-0.075,-0.011]$ & a quarter of the gap to D \\
C$_{\mathrm{half}}$\,$-$\,C (raster four times finer) & 9, \S\ref{sec:raster} & $-0.047$ $[-0.106,+0.001]$ & no resolved gain, localisation worsens \\
D$_{\mathrm{attn}}$\,$-$\,C (learned attention painter, tuned) & 11, \S\ref{sec:setattn} & $+0.235$ $[+0.170,+0.308]$ & worse than no painting at all \\
SetOnly\,$-$\,C (raster-free set decoder, tuned) & 11, \S\ref{sec:setattn} & $+0.219$ $[+0.183,+0.261]$ & worse than the raster core alone \\
\midrule
\multicolumn{4}{l}{\emph{Limits of zero-shot transfer}} \\
\Gtwo{}\,$-$\,C, frozen across the provider boundary & 7, \S\ref{sec:external} & $+0.079$ $[+0.036,+0.121]$ & likelihood lost, localisation kept \\
\Gtwo{}\,$-$\,D$_{\mathrm{raw}}$, frozen & 10, \S\ref{sec:external} & $+0.048$ $[+0.023,+0.070]$ & raw painter ahead; adjusted $p=0.062$ \\
\bottomrule
\end{tabular}
\end{table}

\paragraph{Related work by the authors.}
A companion manuscript, \emph{An Open, Auditable Pass-Surface Benchmark
from Public Bundesliga Tracking Data} (G\"unay and Gun, in preparation),
studies the construction and automated target corroboration of the public
IDSSE benchmark used here. The papers share the upstream data, extraction
implementation and benchmark version. The companion uses a separately
trained reproduction of published SoccerMap, but no \Gtwo{}-family
checkpoint, contrast or result from this paper, and its validation and
sensitivity results are not reported here. The supplement specifies the
external corpus construction independently of the companion, and the cover
letter includes a contribution-overlap disclosure.

\subsection{Research questions and design}

\textbf{RQ1:} Does painting improve endpoint selection and completion
prediction over the same raster-only core across matches and after a change
of provider? \textbf{RQ2:} Which front-end components account for the gain,
and does it persist under per-arm learning-rate selection and the stated
early-stopping rule?

The primary corpus contains 53{,}628 quality-gated passes from all 64 matches
of the 2022 World Cup. The family comprises the raster-only core C, the
raw-feature painter D$_{\mathrm{raw}}$, the per-node encoder D and the
relational encoder \Gtwo{}; controls include the original SoccerMap and a
widened version matched to \Gtwo{}'s parameter count. Cross-validation and
external retraining address RQ1; component ablations, alternative interfaces
and tuned comparisons address RQ2. Thirteen studies recorded their training
designs before their own runs, with the statistical specification and prior
data exposure distinguished in Sec.~\ref{sec:prespec}. The original three-arm
comparison is development history, not independent confirmation.

\subsection{Contributions}
\begin{enumerate}
\item Evidence, under cross-validation over all 64 matches, per-arm
learning-rate selection over five seeds, external retraining and three
surface cores, that continuous-coordinate feature painting improves
endpoint-supervised soccer surfaces over the same raster-only core.
\item A quantified representation trade-off: raw-feature painting carries
three quarters of the gain and learned content the remaining quarter, with
the tested interface alternatives priced against the same core and the
limits of frozen transfer reported.
\item A retained evaluation record: signed study commitments, locked
predictions, match- and seed-level uncertainty, full hypothesis families,
and explicit disclosure of holdout reuse and post-hoc analyses.
\end{enumerate}

Section~\ref{sec:related} describes prior work; Secs.~\ref{sec:data},
\ref{sec:method} and~\ref{sec:protocol} specify the data, models and
evaluation. Section~\ref{sec:results} presents the primary comparisons
before the mechanism studies. Section~\ref{sec:threats} states the limits
and Sec.~\ref{sec:conclusion} concludes. The online supplement (Online
Resource~1) contains the full hypothesis families, construction details and
development records.

\section{Related work}
\label{sec:related}

Three axes locate the contribution: \emph{output support} (per-player
probabilities vs.\ dense pitch cells), \emph{state representation}
(hand-crafted, raster, coordinate sequence, graph) and \emph{temporal
context}. Models with dense spatial outputs read the match state as binned
rasters~\citep{fernandez2020soccermap,fernandez2021epv,overmeer2025revisiting,robberechts2023unxpass},
as sequences of tracking coordinates~\citep{pei2024passing}, or through
hand-specified physics on continuous positions and
velocities~\citep{spearman2017physics}. Relational and graph models keep
players as entities but emit per-player or per-event
outputs~\citep{stockl2021making,rahimian2026tgn,wang2024tacticai}. None of
the soccer systems we reviewed writes learned per-player features onto a
pitch grid at continuous coordinates and trains that encoder end to end
through a dense pitch-cell loss, which is the cell this work occupies.

\paragraph{Dense surfaces.}
SoccerMap~\citep{fernandez2020soccermap} established the three-scale fully
convolutional surface trained by single-pixel supervision and is our
architectural ancestor and control; the EPV framework
of~\citet{fernandez2021epv} supplies the completion/selection/value
decomposition our three heads instantiate. Later work varies that design:
\citet{overmeer2025revisiting} introduce a U-Net-inspired
architecture~\citep{ronneberger2015unet} with a three-level encoder and
attention-gated skips, and add a risk/reward split and an expert-paired
benchmark; our modernised core shares the three-level encoder and the
lateral skips into a top-down path, and adds squeeze-and-excitation and a
dilated bottleneck that theirs does not have, and
Sec.~\ref{sec:raster} reproduces the interface effect on a canonical U-Net
core rather than on their system. The un-xPass system~\citep{robberechts2023unxpass} reuses
SoccerMap-style components to score creativity. \citet{pei2024passing} feed two
seconds of player and ball tracking coordinates to a transformer and predict
the pass-end zone, with separate cross-entropy losses along each pitch axis.
Physics-based pitch control~\citep{spearman2017physics,spearman2018beyond}
is the complementary non-neural family, computed from continuous positions
and velocities. These systems differ in what they read. SoccerMap and its
descendants bin players into raster cells, the sequence model reads
coordinates and decodes zones through attention, and pitch control is
specified by hand. None of them paints learned per-player features onto the
grid at each player's continuous position for a convolutional network to
read, which is the interface this paper evaluates.

\paragraph{Relational models.}
Graph learning for soccer is not new: graph convolution for expected receiver,
pass and threat~\citep{stockl2021making}, possession memory for reception
against defensive structure~\citep{rahimian2026tgn}, and earlier pairwise
spatial relations~\citep{hubacek2019spatial}. These emit distributions over
players or events rather than training a per-cell loss.
TacticAI~\citep{wang2024tacticai} applies geometric deep learning to corner
kicks with graph-level and per-player predictions, but trains no per-cell
surface either. Head-to-head numbers with the closest systems are unavailable
because they share no evaluation universe: \citet{rahimian2026tgn} score
per-receiver predictions on middle-third forward passes from one league, and
\citet{overmeer2025revisiting} use 63 of the same 64 matches as one of two
corpora, alongside 624 Eredivisie matches, under different sourcing and
gating (TRACAB tracking supplied through the Dutch federation, 58{,}569
World Cup passes split for training and evaluation, against our 53{,}628
admitted), so cross-quoting compares pipelines, not architectures. The comparison this
paper can make fairly is the one it makes: against SoccerMap reproductions
trained under an identical protocol.

\paragraph{Entity-to-grid representations and differentiable rasterisation.}
Writing learned per-entity features onto a grid so that a convolutional
network can read them is an established pattern outside sport.
VoxelNet~\citep{zhou2018voxelnet} and PointPillars~\citep{lang2019pointpillars}
encode the points of a LiDAR sweep with a small per-point network and
scatter the result into voxels or pillars for a 2D CNN; the per-point network is a
PointNet~\citep{qi2017pointnet} in both, the pooled-set construction that
Deep Sets~\citep{zaheer2017deepsets} formalises, and attention-based set
encoders~\citep{lee2019settransformer} extend it with relations between
points. SPLATNet~\citep{su2018splatnet} splats point features onto a
permutohedral lattice, and Lift-Splat-Shoot~\citep{philion2020lss} splats
image features into a bird's-eye grid. On the rendering side, spatial
transformer networks~\citep{jaderberg2015stn} made bilinear sampling
differentiable, and soft rasterisers~\citep{kato2018n3mr,liu2019softras} do
the same for the transpose direction, geometry to pixels. Outside computer vision, the encoder and decoder of
GraphCast~\citep{lam2023graphcast} pass messages between a latitude--longitude
grid and a mesh graph, a learned transfer between the two representations
where ours is a fixed write at measured coordinates. These systems differ in where an entity's features land. VoxelNet,
PointPillars and Lift-Splat-Shoot bin each entity to one cell, the last
assigning every point to its nearest pillar before sum pooling, so sub-cell
position is discarded at the write; SPLATNet instead distributes each point
over the surrounding lattice vertices with barycentric weights. Splatting with
fractional weights is itself established machinery, in the bilateral
grid~\citep{chen2007bilateralgrid}, in the bilinear scatter-add of softmax
splatting for forward warping~\citep{niklaus2020softmax} and in
differentiable surface splatting~\citep{yifan2019dss}; the nearest-versus-interpolated
question has been raised inside the pillar literature in the opposite,
gather direction~\citep{wang2020pillar}. Our operator is the bilinear case
of that family written onto a Cartesian grid, with a per-entity encoder that
is Deep-Sets-like in arm~D and relational in \Gtwo{}. What is new is not the
operator but its use: entities are players whose sub-cell position and
own features carry the signal, the grid is a pitch surface supervised at
one observed endpoint per pass, and the paper prices each element of the
front end under a pre-specified multi-seed protocol.

\emph{Graph-to-Grid} is our name for the family of models studied here, not
an established term, and the words are already in use elsewhere.
\citet{lyu2023gpgl} lay general graphs out on a 2-D grid so that a CNN can
read them, and \citet{jin2023g2g} use G2G for a graph-to-grid module that
recodes graph-structured physiological features as a 2-D array, also for a
CNN. The abbreviation is in use for graph-to-graph models as well. In all of
those the grid is a layout to be learned or optimised; here it is the pitch,
every entity is written at its measured physical position, and no graph is
emitted.

\paragraph{Value, evaluation and data validity.}
VAEP~\citep{decroos2019vaep}, pass risk/reward
decomposition~\citep{power2017passes} and line-breaking analyses of the same
tournament~\citep{karakus2025gaps} are reminders that expected possession
value is one of several defensible semantics. \citet{davis2024methodology}
separate model accuracy from indicator validity, and \citet{vanarem2026quality}
quantify such indicators' estimation error against a ground truth football
cannot observe. Broadcast tracking carries provider-dependent position
error~\citep{crang2026validity,mills2026events}, quantified in
Sec.~\ref{sec:threats}. \citet{dwork2015holdout} formalise how sequential
holdout reuse erodes a holdout's guarantees, which is the reason the
attribution ladder was repeated under a pre-specified protocol. Temperature
scaling follows~\citet{guo2017calibration}. Multiplicity control follows
\citet{holm1979} and, for the joint sign-flip test, the max-$T$ construction
of \citet{westfall1993resampling}. Reporting is guided by the checklist
of~\citet{pineau2021reproducibility}.

\section{Data, targets and quality control}
\label{sec:data}

Every pass in the primary corpus is a real pass from the World Cup, with its
destination read off the tracking data where the ball flight resolves to a
tracked next-possession event ($98.9\%$ of admitted passes) and taken from a
geometry-derived boundary exit or a receiver-capture rule for the remainder
(Sec.~\ref{sec:endpoints}).

\subsection{Corpus, loader and split}

\begin{figure}[tbp]
\centering
\includegraphics[width=0.92\textwidth]{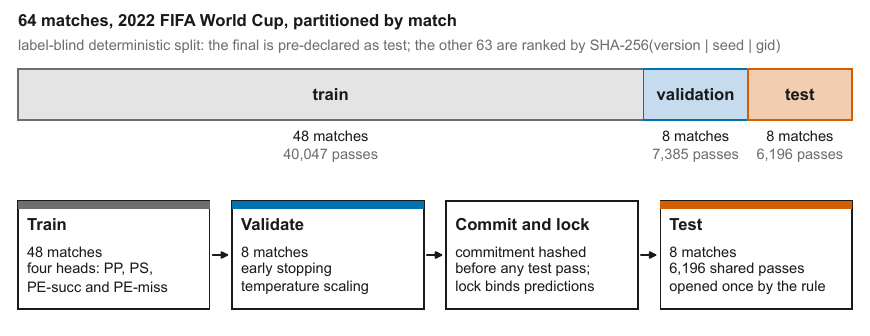}
\caption{The shared 48/8/8 match-grouped protocol: a label-blind
deterministic split, validation-only early stopping and temperature scaling,
and a hash-committed test evaluation of 6{,}196 shared passes, which the
protocol opens once per arm. In the realised runs the \Gtwo{} arm opened
them twice, the first cycle aborted by an automated integrity failure and
each control opening once; the records are in Sec.~\ref{sec:records} and
Sec.~\ref{sec:threats}}
\label{fig:protocol}
\end{figure}

We use the PFF FC release~\citep{pff2024worldcup} covering all 64 matches of
the 2022 FIFA World Cup: event JSON, metadata and broadcast tracking; our copy
stores smoothed tracking at 29.97\,Hz, about 50\,GB and on the order of $10^7$
frames. It gives full-tournament coverage without instrumented stadiums, at
the cost of players leaving frame, intermittent ball visibility and smoothing
artifacts, so the pipeline treats data quality as a modeling problem. The
corpus contains \textbf{53{,}628} admitted passes at $83.8\%$ completion,
split 48/8/8 by match into 40{,}047 training, 7{,}385 validation and 6{,}196
test rows (Fig.~\ref{fig:protocol}).

All raw access flows through one self-verifying loader whose audited source
modules are embedded as SHA-256-checked payloads; any modification fails the
import. A ball-in-play state machine closes play at out-of-play, period end
and awarded fouls and reopens at set-piece releases or first open-play
control. We additionally mask the 5\,s after free kicks, corners, throw-ins
and penalties, removing set-piece routines from an open-play pass model.
Quality intervals flag illegal roster sizes, observational blackouts, provider
dead time, frozen scenes, inter-frame gaps and impossible speeds, and a pass
snapshot is admitted only under joint constraints on event--frame latency,
roster completeness at the snapshot and at the 1\,s-earlier frame used for
finite-difference velocities, actor--ball distance and tracking-run
continuity. Coordinates are rotated per possession so the acting team always
attacks toward $x=105$.

Matches are split 48/8/8 by a label-blind deterministic rule: the final is
pre-declared as a test anchor and the remaining 63 are ranked by
$\mathrm{SHA256}(\text{version}\,|\,\text{seed}\,|\,\text{gid})$ with seed
1907, so no outcome or model output can influence membership. A 192-file
SHA-256 manifest binds the raw data to the split signature
\texttt{16cf1b7a...}, identical across arms.

\subsection{Tracking-grounded endpoints}
\label{sec:endpoints}

Where a pass \emph{ended} is the label the selection model is scored against,
so we take it from the tracking rather than from annotation. Each eligible
pass opens a 6\,s ball-flight window resolved to the tracked ball position at
the next possession event (53{,}041 passes, $98.9\%$), a geometry-derived
pitch-boundary intersection (476, $0.89\%$), or the first frame where the ball
comes within 1\,m of the annotated receiver after $\geq$2\,m of travel (111,
$0.21\%$); unresolved flights are quarantined. Where we stratify by provenance
the reported strata are next-event ($6{,}135$ rows) and boundary geometry
($52$); the $9$ receiver-capture rows are too few to bootstrap and are omitted
there, though present in every pooled number. Admission is not outcome-blind
in effect even though it is outcome-blind by rule: completion rates are
$83.8\%$, $85.3\%$ and $81.9\%$ in the training, validation and test splits,
and $6.6\%$, $5.9\%$ and $7.4\%$ of passes carry a nonzero value label. The
loader logs every gate; the per-gate exclusion count on this corpus
(68{,}566 candidate pass events, 11{,}003 removed by the ball-in-play
contract, 3{,}935 by the snapshot and endpoint gates, 53{,}628 admitted) and the admission rates by pass type are given in Supplement
Sec.~\suppref{supp:gates}, Tables~\suppref{tab:gate_audit}
and~\suppref{tab:gate_audit_bytype}. A validation of the endpoint rule against
manual annotation is not part of this paper (Sec.~\ref{sec:threats}). The
public seven-match corpus that Sec.~\ref{sec:external} uses as a second
provider is built by the same rules; its construction is specified in full
in Supplement Sec.~\suppref{supp:external}.

\subsection{Targets}
\label{sec:targets}

Let $k^{*}$ be the grid cell of the observed endpoint on the $104\times68$
grid ($\approx$1\,m cells, 7{,}072 cells). \textbf{PP} (pass probability) is
trained with binary cross-entropy between the completion label $y\in\{0,1\}$
and $\sigma(z_{k^{*}})$, the sigmoid of the logit at the observed endpoint
cell; \textbf{PS} (pass selection) as a 7{,}072-way softmax with $k^{*}$ as
the target class. \textbf{PE} (pass value) regresses a signed 15-second
expected-goal balance,
\begin{equation}
r=\Bigl[1-\!\!\prod_{s\in S_{\mathrm{own}}}\!\!(1-x_s)\Bigr]
 -\Bigl[1-\!\!\prod_{s\in S_{\mathrm{opp}}}\!\!(1-x_s)\Bigr]\in[-1,1],
\label{eq:r15}
\end{equation}
over shots by each team within 15\,s of the pass, $x_s$ being an
expected-goal value clipped at $0.999$. The xG estimator is \emph{external} to
the tournament: an XGBoost classifier~\citep{chen2016xgboost} fitted on the
80{,}304-shot training partition of a 4{,}171-match, 106{,}787-shot StatsBomb
open-data corpus~\citep{statsbomb_opendata} from which the 2022 World Cup was
removed before cache construction (match-grouped $75\%$ split; 26{,}483-shot
diagnostic holdout, log-loss $0.285$). Two heads condition on outcome,
$\PE_{\mathrm{succ}}$ on completed and $\PE_{\mathrm{miss}}$ on missed passes,
each regressing $\tilde r=(r+1)/2$ of Eq.~\eqref{eq:r15} under MSE at $k^{*}$;
predictions map back by $\hat v=2\sigma(z)-1$ onto the signed $[-1,1]$ scale.
The combined surface is
\begin{equation}
\widehat V(k)=\hat p(k)\,\hat v_{\mathrm{succ}}(k)+\bigl(1-\hat p(k)\bigr)\hat v_{\mathrm{miss}}(k).
\label{eq:epv}
\end{equation}
Because every head is supervised at a single observed cell $k^{*}$ per pass,
the surfaces away from $k^{*}$ are extrapolations of the network, a point
Sec.~\ref{sec:threats} returns to under calibration.

\section{Method}
\label{sec:method}

\begin{figure}[t]
\centering
\includegraphics[width=\textwidth]{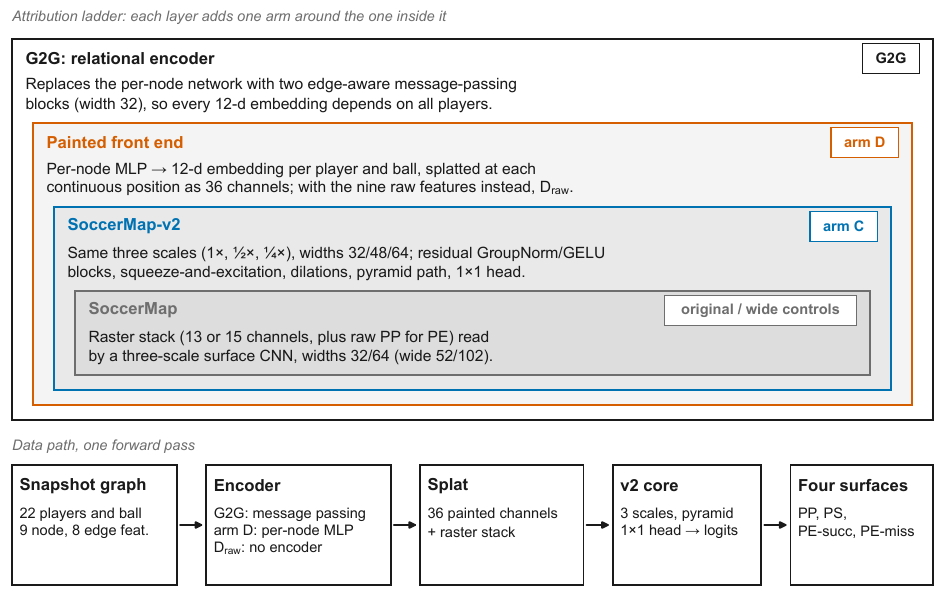}
\caption{\Gtwo{} as three layers around the original SoccerMap, each layer one
arm of the attribution ladder. Innermost, \emph{SoccerMap}: raster channels
read by a three-scale surface CNN (the original and wide controls).
\emph{\vtwo{}} (arm~C) keeps that layout with residual GroupNorm/GELU blocks,
squeeze-and-excitation, dilations and a pyramid path. The \emph{painted front
end} (arm~D) describes every player and the ball in 12 numbers and splats them
onto the grid at each one's continuous position as 36 extra channels.
\emph{\Gtwo{}} replaces that per-node network with two edge-aware
message-passing blocks; Sec.~\ref{sec:attribution} prices that replacement; with the nine raw node features painted in place of the embedding, the front end is D$_{\text{raw}}$.
Below, the data path}
\label{fig:arch}
\end{figure}

\Gtwo{} describes each player, paints that description onto the pitch where
the player stands, and reads the painted pitch with a convolutional network.
It is built as three layers around the original SoccerMap
(Fig.~\ref{fig:arch}), and each layer is one arm of the experiments.

\subsection{What is new, and what is inherited}

The interface is established; the contribution is its evaluation.
A differentiable splat writes entity features into a grid, and endpoint
supervision reaches any encoder that produced them. The study compares
bilinear and nearest-cell placement, raw and learned features, and raster and
attention alternatives. Fractional-weight splatting, the SoccerMap architecture, and the
completion/selection/value decomposition are inherited
(Sec.~\ref{sec:related}). Raw painting is the simpler recommended baseline;
the learned G2G front end has the better in-distribution selection likelihood
in the primary comparisons.

The system nests. \textbf{SoccerMap} is the base: raster channels in, a
three-scale fully convolutional network, a logit surface out.
\textbf{\vtwo{}}, our name for the modernised core and not a release by the
SoccerMap authors, keeps that layout and input but modernises every block
(Sec.~\ref{sec:core}); alone, it is arm~C. The \textbf{painted front end}
adds a per-node network and the splat (Sec.~\ref{sec:splat}), 36 learned
channels alongside the raster stack; with it, arm~D. \textbf{\Gtwo{}}
replaces the per-node network with the relational encoder
(Sec.~\ref{sec:encoder}). The chain
$\text{SoccerMap}\to\text{v2}\to\text{D}\to\Gtwo$ is also the attribution
ladder of Sec.~\ref{sec:attribution}; the arms are specified, with parameter
counts, in Sec.~\ref{sec:arms}.

\subsection{\vtwo{}: the surface core}
\label{sec:core}

The core reads the raster stack
of~\citet{fernandez2020soccermap,fernandez2021epv}: 13 channels for PP/PS and
15 for PE, plus a 16th carrying the model's own raw PP surface
(Sec.~\ref{sec:stacking}). In the arms that paint, the 36 painted channels are
concatenated to it, giving input widths $49$ and $52$. The core keeps
SoccerMap's three scales, processing at $1\times$, $\tfrac12\times$,
$\tfrac14\times$, and differs in the blocks. Widths become $(32,48,64)$. Each scale's pair of plain
$5\times5$ convolutions becomes a residual~\citep{he2016resnet} block with
GroupNorm~\citep{wu2018groupnorm} and GELU~\citep{hendrycks2016gelu}, followed
by squeeze-and-excitation~\citep{hu2018senet}; the coarsest scale adds dilated
convolutions. SoccerMap's per-scale predictions with nonlinear upsampling and
concatenation fusion become a top-down pyramid path~\citep{lin2017fpn} that
restores full resolution before a single $1\times1$ head emits the logit
surface. PP, PS, PE-succ and PE-miss are independently trained instances of
the whole network; no weights are shared across heads or arms.

\subsection{The painted front end: splatting}
\label{sec:splat}

Each player and the ball is a node with nine features --- normalised position,
velocity, team and ball-role identity --- on fixed physical scales, nothing
fitted to the corpus. A per-node network maps those features to a
12-dimensional embedding $e_i\in\mathbb{R}^{12}$, and the splat writes that
embedding into the surface domain at the node's \emph{continuous} grid
coordinates. The features can also be painted directly, with no network, and
Sec.~\ref{sec:ablation} compares them with learned per-node embeddings. Let $(u_i,v_i)$ be node $i$'s position in cell units, let
$(\bar u_i,\bar v_i)=(\lfloor u_i\rfloor,\lfloor v_i\rfloor)$ and
$(\delta_i,\epsilon_i)=(u_i-\bar u_i,\,v_i-\bar v_i)$. For each of the three
groups $g\in\{\text{attackers},\text{defenders},\text{ball}\}$ the painted
channel block $S_g\in\mathbb{R}^{12\times68\times104}$ is
\begin{align}
S_g[c,y,x]&=\sum_{i\in g}\;\sum_{a,b\in\{0,1\}}
 w_{ab}(\delta_i,\epsilon_i)\,
 \mathbb{1}[x=\bar u_i+a]\,\mathbb{1}[y=\bar v_i+b]\;e_i[c],
\label{eq:splat}\\
w_{ab}(\delta,\epsilon)&=\delta^{a}(1-\delta)^{1-a}\,\epsilon^{b}(1-\epsilon)^{1-b},
\nonumber
\end{align}
The implementation uses cell-centre coordinates
\[
u_i=\operatorname{clip}(W x_i/105-1/2,0,W-1),\qquad
v_i=\operatorname{clip}(H y_i/68-1/2,0,H-1)
\] for physical positions
$(x_i,y_i)$, with $W=104$ and $H=68$. The upper neighbours are clipped to
the last valid indices. Coincident boundary contributions accumulate, so the
weights still sum to one; clipping sacrifices positional resolution in the
outer half-cell strips. Masked nodes contribute zero. An implementation audit
checks signed-feature conservation, permutation invariance, boundary cases,
and embedding and interior-coordinate gradients
(Supplement Sec.~\suppref{supp:compute}). Equation~\eqref{eq:splat} is
therefore bilinear weights over the four surrounding cell centres accumulated
by scatter-add, $3\times12=36$ channels in all. This is the painting step. It is linear in the embeddings and its Jacobian with respect to $e_i$ is the
weight pattern itself, so the surface loss backpropagates through the grid
into whatever network produced the embedding. Positions are inputs here, and
the map is differentiable in them almost everywhere, with kinks at cell
boundaries. Three properties matter for what follows. First, a node sitting a
fraction $\delta$ inside a cell contributes $1-\delta$ of its embedding to
that cell and $\delta$ to the next, so the painted channels carry sub-cell
position up to bilinear interpolation, whereas an occupancy raster carries
only the cell. When two nodes of one group fall in the same cell their
contributions add. Second, the operator is permutation-invariant within a group and
defined for any number of nodes (its output scales with that number, fixed
here at 22 players plus the ball up to players out of frame). Third,
nothing in the operator is fitted: every parameter of the front end lives
in the embedding network. With a per-node MLP as that network, the front end is arm~D: nodes
are described independently and never exchange information. With the nine
raw node features in place of the embedding, it is D$_{\text{raw}}$, which
fits no front-end parameter at all (Sec.~\ref{sec:ablation}).

\subsection{The relational encoder}
\label{sec:encoder}

\Gtwo{} replaces the per-node MLP with an encoder in which players see each
other. The snapshot becomes a fully connected graph over up to 23 nodes (up to
11 per side plus the ball) with eight directed edge features of pairwise
geometry. Two residual message-passing blocks of hidden width 32 follow the
graph-network formalism~\citep{battaglia2018relational}. The attention logits
combine scaled dot-product content similarity~\citep{vaswani2017attention}
with an edge-geometry bias, following attention over graph
neighbourhoods~\citep{velickovic2018gat,brody2022gatv2}:
\begin{equation}
\alpha_{ij}=\operatorname*{softmax}_{j\in\mathcal N(i)}
\Bigl(\tfrac{\langle q_i,k_j\rangle}{\sqrt{32}}+b(e_{ij})\Bigr),
\label{eq:attention}
\end{equation}
with $b$ a two-layer MLP on edge features. Messages
$m_{ij}=\mathrm{MLP}_m([h_j;e_{ij}])$ aggregate as
$a_i=\sum_j\alpha_{ij}m_{ij}$ under a residual layer-normed update, and a
linear layer emits the 12-dimensional embedding per node that the splat
paints. The encoder is permutation-equivariant and anonymous: no player
identity is used. Because the splat of Eq.~\eqref{eq:splat} is linear in $e_i$ and the attention
of Eq.~\eqref{eq:attention} is smooth in its inputs, the whole chain from node
features through message passing to the logit surface is differentiable. The
endpoint supervision of Sec.~\ref{sec:targets} is the only supervision the
encoder ever receives.

\subsection{Estimands, supervision and what the surfaces mean}
\label{sec:estimands}

Every head produces a dense output surface but is supervised sparsely, at the
one observed endpoint per pass, and the three heads estimate different
things. \textbf{PP} estimates the probability that a pass played to a cell is
completed, \emph{conditional on that cell being the observed endpoint}. It is
trained with binary cross-entropy at $k^{*}$ only, so its values at cells the
passer did not choose are extrapolations by the network from historically
chosen endpoints, and no label ever tests them. \textbf{PS} estimates the
distribution of the observed next endpoint over the 7{,}072 cells: its
softmax gradient touches every logit, so PS is the one head whose whole
surface is shaped directly by the loss. \textbf{PE} estimates the expected value of the
signed 15-second expected-goal balance, conditional on outcome and endpoint,
again supervised at $k^{*}$ only. The endpoint is where the ball was next observed, not where the passer
intended it to go (Sec.~\ref{sec:threats}). Away from observed endpoints the
convolutional core supplies only its locality and weight-sharing biases, not
an explicit smoothness constraint. This is the standard SoccerMap arrangement,
and it is the reason we report both target-pixel metrics and, for selection,
the full distribution's mass and rank statistics. Wherever this paper says a surface is \emph{calibrated}, it
means temperature-scaled at observed endpoints; calibration away from them
is not established (Sec.~\ref{sec:threats}). Optimisation, early stopping,
calibration and the leak-free stacking of the PP surface into the PE heads
are shared by every arm and are specified in Sec.~\ref{sec:stacking}.

\subsection{Implementation}

The models are implemented in PyTorch. The shipped arms trained from the
original notebooks, and the multi-seed studies from a standalone training
engine that reproduces the shipped predictions to a maximum absolute
probability difference of $1.63\times10^{-4}$. AI assistance with code and
text is described in the Statements and Declarations.

\section{Experimental protocol}
\label{sec:protocol}

Every arm sees the same matches, the same passes, the same training allowance
and the same scoring code, and each evaluation opens the held-out test matches
(the \emph{holdout}) only after the models are frozen. This section specifies the arms, the shared
training recipe, the evaluation records, the inference machinery and the
thirteen pre-specified studies; the supplement gives each study's design in
full.

\subsection{Arms}
\label{sec:arms}

\textbf{Original SoccerMap} is a topology-faithful reproduction
of~\citet{fernandez2020soccermap}: three scales, two $5\times5$
convolution+ReLU blocks per scale at widths 32/64, $1\times1$ prediction at
every scale, nonlinear upsampling and concatenation fusion, consuming the same
deterministic channels and no graph channels. It totals $274{,}251$
parameters; the published architecture leaves some widths unspecified, so all
control counts here refer to our instantiations. \textbf{Wide SoccerMap} is
the identical topology at widths 52/102, chosen so its parameter count matches
\Gtwo{}: $698{,}573$ vs.\ $700{,}353$ for PP/PS, a difference of $1{,}780$
parameters ($0.25\%$), and $702{,}473$ vs.\ $702{,}753$ for PE ($0.04\%$). No
advantage of \Gtwo{} can be attributed to parameter count alone.

Two further arms split the front end. \emph{Arm~C} is \vtwo{} alone, fed only
the deterministic channels, with the painting step and node encoder deleted
together ($654{,}899$ PP/PS parameters). \emph{Arm~D} restores the painting
step and a per-node MLP but removes both message-passing blocks, so nodes are
described independently and never exchange information ($684{,}479$; its
projector holds $780$ of \Gtwo{}'s $16{,}654$ projector parameters). The chain
$\text{wide}\to\text{C}\to\text{D}\to\Gtwo$ isolates, in order, \vtwo{}, the
painting interface and message passing. The graph branch's full marginal cost over arm~C is $45{,}454$ parameters:
$16{,}654$ in the projector plus $28{,}800$ that the 36 painted channels add
to the core's first convolution. That is a $6.9\%$ increase over \vtwo{} and
$6.5\%$ of the final model.
Table~\ref{tab:arms} lists every arm with both head groups' counts.

\begin{table}[t]
\caption{The three shipped arms (top) and the three attribution arms (bottom); D$_{\text{raw}}$ trains PP and PS heads only.
All share data, targets, loader, split, optimiser, losses, calibration,
metrics and evaluation code. The convergence reruns of Supplement
Sec.~\suppref{sec:convergence} retrain only the two controls' PS heads with
the epoch cap raised to 200 and are otherwise identical to their shipped
counterparts. Parameter counts refer to our instantiations; the published
SoccerMap leaves some widths unspecified}
\label{tab:arms}
\centering
\small
\begin{tabular}{lrrl}
\toprule
Arm & PP/PS parameters & PE parameters & Input / system \\
\midrule
\Gtwo{}              & 700{,}353 & 702{,}753 & graph + splat + raster \\
Original SoccerMap   & 274{,}251 & 276{,}651 & raster only \\
Wide SoccerMap       & 698{,}573 & 702{,}473 & raster only \\
\midrule
Arm D (splat, no message passing) & 684{,}479 & 686{,}879 & splat + raster \\
Arm D$_{\text{raw}}$ (raw features painted, no encoder) & 676{,}499 & --- & splat + raster \\
Arm C (\vtwo{} alone)             & 654{,}899 & 657{,}299 & raster only \\
\bottomrule
\end{tabular}
\end{table}

\subsection{Training, calibration and stacking}
Here ``tuned'' means learning-rate selection on the stated grid, not exhaustive
hyperparameter optimisation. ``Converged'' is shorthand for satisfying the
stated validation early-stopping rule before the epoch cap; it is not a
claim of numerical or global optimisation convergence.

\label{sec:stacking}

All shipped arms trained with Adam under bf16 autocast on a single RTX 5090
(mean logged epoch $42.6$\,s for \Gtwo{} against $19.8$\,s and $20.5$\,s for
the controls; training windows $0.917$, $1.379$ and $0.912$\,h). The
pre-specified studies trained on an NVIDIA RTX PRO 6000 Blackwell workstation
GPU, the attribution ladder among them in under four GPU hours. Every arm
shares per-head learning rates (PP $10^{-4}$, PS $10^{-5}$, PE $10^{-6}$),
batch sizes, recorded seeds (which in the shipped runs governed data order
only; Sec.~\ref{sec:threats}), and early stopping on the validation objective
(patience 5, cap 50 epochs, an epoch being one full sweep through the training
data). We call this shared setting the \emph{recipe}; Supplement
Sec.~\suppref{app:repro} tabulates every input feature, hyperparameter and
selected epoch and temperature. After weights freeze, one scalar temperature
per probabilistic head is fitted \emph{on validation only}: a single number
improving agreement between predicted percentages and observed frequencies,
binary for PP, spatial for PS~\citep{guo2017calibration}. PE is left
uncalibrated. Because PP is trained and scored at the historically
chosen destination cell, this is \emph{target-pixel} calibration.

The PE heads consume the arm's own raw PP surface as an input channel. To
prevent a producer from scoring a match it trained on, the training matches
are partitioned into four interleaved folds and a dedicated PP producer
predicts only its held-out fold. Validation and test rows receive surfaces
from the main PP model, which never trained on them; a 53{,}628-row provenance
table records the producer of every row.

\subsection{Evaluation records}
\label{sec:records}

Before any test forward pass, each evaluation cycle writes a hash-verified
\emph{commitment} freezing the test match list, checkpoint hashes, both
temperatures, a metric menu and the hash of eight pre-declared scenes. A later
\emph{lock} binds it to the prediction-file hashes. The shipped arms were
evaluated on 14, 16 and 17 August 2026 on the same eight matches. The \Gtwo{}
arm's log records two commitment/evaluation cycles, the first aborted by an
automated integrity failure; Sec.~\ref{sec:threats} discusses the consequence
and Supplement Sec.~\suppref{app:firstcycle} prints the discarded cycle in
full.

\subsection{Paired inference}
\label{sec:paired}

\paragraph{Estimands.}
The eight test matches are the resampling units: passes within a match are
not independent, teams recur across matches, and the final was anchored by
rule rather than drawn. The \emph{pass-weighted} estimand is the difference in a metric between two
arms over the population of passes from which the 6{,}196 test passes are a
sample; it is estimated by the pooled difference over all test rows. The
\emph{equal-match} estimand is the expected per-match difference over the
population of matches; it is estimated by the mean of the eight per-match
differences, which weights the final and a group-stage match equally. Both are reported for every contrast (Supplement
Table~\suppref{tab:paired-full}), together with a leave-final-out pooled
estimate for the primary metrics.

\paragraph{Interval estimation.}
Because all arms are scored on identical rows, the comparative question is
answered by resampling the eight test matches with replacement ($B=10{,}000$,
seed 1907) and recomputing $\Delta=\Gtwo-\text{control}$ on the same
resampled rows in each replicate. We report percentile match-cluster
bootstrap intervals~\citep{efron1994bootstrap} obtained from those 10{,}000
resamples; bias-corrected and accelerated intervals agree on the exclusion
of zero in every case. The 13-metric \emph{family}, the set of hypotheses over which multiplicity is
controlled, spans PP log-loss, Brier, AUC and ECE, and five PS metrics: NLL,
mean argmax distance, probability mass within 3\,m and 5\,m of the observed
endpoint, and the rank fraction of the observed cell. It also spans both
PE-branch MSEs and combined EPV MSE and MAE.

\paragraph{Sign-flip inference and its scope.}
With eight clusters, bootstrap tail proportions are floored at $1/B$ and are
not valid confirmatory $p$-values: few-cluster cluster bootstraps
over-reject~\citep{cameron2008clustered}, and an uncentred tail proportion is
not a test of a null. We therefore prioritise the joint sign-flip max-$T$ analysis over those
bootstrap tail probabilities. We studentise the eight match-level
paired deltas, $t=\mathrm{mean}(d)/(\mathrm{sd}(d)/\sqrt{8})$, and recompute
$t$ under all $2^{8}=256$ cluster sign-flip patterns; the single-step max-$T$
adjusted $p$~\citep{westfall1993resampling} for comparison $j$ is the
fraction of patterns whose family-wide maximum $|t|$ reaches the observed
$|t_j|$. Its error control assumes that the paired match-level deltas are
exchangeable in sign under the joint null, an assumption that match grouping
motivates but does not guarantee (Sec.~\ref{sec:threats}); a global flip leaves $|t|$ unchanged, so the
128 mirror pairs make $2/256=0.0078$ the smallest attainable adjusted $p$,
still below $\alpha$. The statistic tests the equal-match estimand, so it is
the natural companion of the equal-match column in every table. For the
multi-seed studies the same test is applied to the seed-averaged per-match
differences, which conditions on the seeds drawn; the hierarchical bootstrap
of Sec.~\ref{sec:prespec} supplies the intervals that carry seed
variability. For comparability with the earlier analysis, Holm adjustment~\citep{holm1979}
of the bootstrap $p$-values is also reported in every full-family table of the
supplement. It is applied \emph{post hoc} across the 26 reported contrasts of
this retrospective comparison (the shipped commitment fixed a broader metric
menu, and this $13\times2$ family and its correction were settled after the
locked predictions existed) and within the pre-declared family of each later
study. Where the two \emph{procedures}, max-$T$ and Holm, disagree, the
max-$T$ verdict is the one the paper reports. Two unadjusted per-comparison checks of the shipped family, the exact
sign-flip and Wilcoxon tests, are given in Supplement
Sec.~\suppref{app:signtests}.

\subsection{Pre-specified multi-seed studies}
\label{sec:prespec}

Every study after the shipped comparison was run under a stricter
discipline than that comparison, and all share an auditable workflow, with study-specific families and seed counts. Each writes a self-hashed \emph{commitment record} before any of its runs
begins training. The record fixes the run matrix (arm $\times$ seed, with
every fit seed derived from a recorded seed base by a fixed formula), the
study-specific metric family (14 metrics for full four-head studies, and
10 PP/PS metrics where value is not evaluated), two primary endpoints
(calibrated PP log-loss and calibrated PS NLL), the comparison pairs, the Holm
family and $\alpha=0.05$, and the bootstrap design. It also fixes a test-set
policy: no test label is read before a \emph{lock record} has hashed every
checkpoint of every run.
Training touches train and validation rows only; the one array written
during training that spans test rows is the PE raw-PP input channel, which is
forward inference of the run's own PP model and consumes no label.
Evaluation refuses to train and fails if any checkpoint is missing or no
longer matches its locked digest. The commitment also records the SHA-256 of every code file involved, and the
later phases refuse to proceed under code drift. One study needed an override: in Study 3 an analysis-arm naming rule was
repaired after the lock and after the single test opening, before any analysis
output existed, and the analysis was re-run with the drift recorded in the run
status; the training runs, the locked predictions, the metric family, the pairs
and the bootstrap design are as committed.

A pre-submission audit found two further defects, both in the analysis code of
the two cross-validation studies. In Studies 5 and 12 Part~A the sign-flip
max-$T$ divided every sign-flipped cluster mean by the standard error of the
observed deltas instead of recomputing it under each pattern, as this section
specifies and as the other eleven studies implement; and the AUC and
calibration-error intervals resampled a row-weighted mean of per-match values
while the printed estimate is the metric on the pooled predictions. Both
families were recomputed from the same locked predictions, under the committed
bootstrap draws and sign patterns, with the studentised statistic described
above and the exact pooled resampling used elsewhere in the paper. The point
estimates are unchanged, no rejection is lost, nine further hypotheses across
the two families newly clear max-$T$, and no interval changes whether it
excludes zero. The correction can only lower these $p$-values: dividing each
hypothesis by a constant leaves a valid but under-powered variant of the same
single-step test, so no rejection reported here depends on the change having
gone in our favour. One reading is strengthened by it, the learned-content
increment of Sec.~\ref{sec:cv}, which now clears both procedures on every
metric of its family except calibration error. The uncorrected families are
retained beside the corrected ones.

The analysis is a \emph{hierarchical bootstrap}, described here for the
three-seed holdout studies; the tuned studies draw five seeds, the
cross-validation resamples 64 match clusters with one seed per fold, and the
external studies resample seven. Each of $B=10{,}000$ replicates draws the 8
test matches with replacement \emph{and} the 3 seeds with replacement,
independently; each arm's metric is the mean over the drawn
seeds evaluated on the drawn rows, and the pair delta is $A-B$ under that same
draw. Its intervals therefore carry match and initialisation variability
together. Two-sided $p$-values are $2\min\{P(\Delta\le0),P(\Delta\ge0)\}$,
floored at $1/B$, and Holm-adjusted over the pooled family. Every bootstrap statistic is computed from per-match sufficient statistics, so
a resample is an exact contraction rather than a re-scan of the 6{,}196 rows.
The fixed-bin ECE and the Mann--Whitney AUC decompose exactly, and the
analysis script re-derives a sample of replicates the naive way and asserts
equality.

One engine amendment applies to every one of the pre-specified studies and matters for the
reproducibility note in Sec.~\ref{sec:threats}: seeding now runs \emph{before}
model construction, so a recorded seed fixes weight initialisation as well as
data order. Replicate $r\in\{1,2,3\}$ uses seed base $1907+100{,}000\,r$; seed
base $1907$ reproduces the shipped runs' data order.

Table~\ref{tab:studymap} lists the thirteen studies in the order in which they
are introduced here, with the archive folder that holds each one's commitment
record, lock record, runs and analysis. Supplement
Sec.~\suppref{supp:designs} gives every study's design as it was committed:
arms, seeds, comparison pairs, family, bootstrap, and the time and signature
of the commitment; each results subsection restates the design of its own
study. Studies that reuse reference predictions pin their digests in the
commitment; the cross-validation, tuned, and external-retraining protocols
instead fit the arms specified for those protocols. Study 4 and Study 11 are adaptive by design and pre-specified as such. A
validation-only sweep on seed 1 trains each head at every rate of a committed
grid to a cap of 200 epochs. A committed rule selects the rate with the lowest
best-validation loss and signs the selection. Seeds 2--5 train at the chosen
rates, and the test set is opened once after every checkpoint is locked.

Study 12 takes the raw-feature painted arm D$_{\text{raw}}$ of Study 3
($676{,}499$ parameters: the 13 raster channels plus the 27 painted raw node
features, no learned encoder) through both primary protocols, in two parts
with separate commitment records, each signed before any run of its part
began training. Part~A trains it in the eight folds of Study 5 at the
shipped recipe, one seed per fold with the Study 5 seed bases and fold table,
and compares its pooled out-of-fold predictions with the locked Study 5
out-of-fold rows of C and \Gtwo{}: two pairs (D$_{\text{raw}}-$C,
\Gtwo{}$-$D$_{\text{raw}}$), a family of 20, the 64-cluster bootstrap and
the Monte-Carlo max-$T$ test of Study 5. Part~B takes it through the
Study 4 protocol, a validation-only sweep on seed 1 over the Study 4 grid,
the committed selection rule, seeds 2--5 at the chosen rates, cap 200,
against the locked Study 4 tuned runs of C and \Gtwo{}: two pairs, a
family of 20, a hierarchical bootstrap over 8 matches and 5 seeds and the
exact 256-pattern max-$T$ test. Neither reference arm is retrained in
either part.

Study 13 asks whether the interface effect is a property of the painting
step rather than of the \vtwo{} core. Two arms are built on a canonical
three-level 2-D U-Net at widths 32, 72 and 160, the triple closest to the
doubling pattern among the 31 of the $7{,}564$ enumerated width triples
that land within 1\% of arm~C's parameter count: U$_{\text{raster}}$ reads
the 13 raster channels ($652{,}833$ parameters) and U$_{\text{raw}}$ the
same channels plus the 27 painted raw node features of D$_{\text{raw}}$
($660{,}609$; the $7{,}776$ extra parameters all in its first
convolution). Each trains PP and PS heads at the shipped recipe under the
three Study 1 seeds and is compared with the locked runs of C and \Gtwo{}
(Study 1) and D$_{\text{raw}}$ (Study 3), never retrained: four pairs
(U$_{\text{raw}}-$U$_{\text{raster}}$, U$_{\text{raster}}-$C,
U$_{\text{raw}}-$D$_{\text{raw}}$, U$_{\text{raw}}-$\Gtwo{}), a family of
40, a hierarchical bootstrap over 8 matches and 3 seeds, the exact
256-pattern max-$T$ test; its commitment was signed before training.

\begin{table}[t]
\caption{The thirteen pre-specified studies. \emph{Study} is the name used in this paper; \emph{archive} is the study's original name, under which the study archive holds its commitment record, lock record, runs and analysis (folder \texttt{overnightN}, lower case, no hyphen; the archive keeps its original names); \emph{commitment} is the first eight hex digits of the SHA-256 signature of the self-hashed commitment record written before any run of the study began training. LOMO: leave-one-match-out. The shared design, the per-study families and every deviation are given in Supplement Sec.~\suppref{supp:designs}}
\label{tab:studymap}
\centering
\scriptsize
\setlength{\tabcolsep}{3.5pt}
\begin{tabular}{clllp{40mm}p{48mm}}
\toprule
Study & Archive & Sec. & Commitment & Arms $\times$ seeds & Tests \\
\midrule
 1 & Overnight-2 & \ref{sec:attribution} & \texttt{1f549421} & \Gtwo{}, C, D $\times$ 3 seeds; cap-200 PS reruns of both controls & attribution ladder: front end, painted package, message passing \\
 2 & Overnight-3 & \ref{sec:history} & \texttt{7522b56a} & original, wide $\times$ 3 seeds vs.\ locked \Gtwo{} & shipped comparison with seed variance on both sides \\
 3 & Overnight-5 & \ref{sec:ablation} & \texttt{0bf9037a} & D$_{\text{nearest}}$, D$_{\text{gauss}}$, D$_{\text{raw}}$, C$_{\text{matched}}$, \Gtwo{}$_{\text{nearest}}$ $\times$ 3 & painting geometry, anti-aliasing, learned vs.\ raw features, parameter count \\
 4 & Overnight-4 & \ref{sec:tuned} & \texttt{07be9096} & \Gtwo{}, C, original, wide $\times$ 5 seeds, tuned & per-arm learning rate and convergence \\
 5 & Overnight-6 & \ref{sec:cv} & \texttt{3789298f} & 4 arms $\times$ 8 folds, all 64 matches & the effect over the whole tournament \\
 6 & Overnight-7 & \ref{sec:setattn} & \texttt{5c107078} & D$_{\text{attn}}$, SetOnly $\times$ 3 & attention painter; raster-free set decoder \\
 7 & Overnight-8 & \ref{sec:external} & \texttt{4ff4c7ab} & 4 arms: zero-shot (3 seeds) and 7 LOMO folds on IDSSE & transfer and replication on a second provider \\
 8 & Overnight-9a & \ref{sec:raster} & \texttt{5cef6d46} & C$_{\text{offsets}}$, original$_{\text{splat}}$ $\times$ 3 & sub-cell offsets on the raster; the splat on the original core \\
 9 & Overnight-9b & \ref{sec:raster} & \texttt{6629aeb1} & C$_{\text{half}}$, original$_{\text{half}}$ $\times$ 3 & a raster four times finer \\
 10 & Overnight-10a & \ref{sec:external} & \texttt{341c1f84} & D$_{\text{raw}}$: zero-shot (3 seeds) and 7 LOMO folds on IDSSE & the raw-feature painter across the provider boundary \\
 11 & Overnight-10b & \ref{sec:setattn} & \texttt{faafcd51} & D$_{\text{attn}}$, SetOnly $\times$ 5 seeds, tuned & the interface alternatives at their own learning rates \\
 12 & Overnight-11 & \ref{sec:cv} & \texttt{261af843} & D$_{\text{raw}}$: 8 CV folds (A); tuned $\times$ 5 seeds (B) & the recommended variant under the primary protocols \\
 13 & Overnight-12 & \ref{sec:raster} & \texttt{3327ef11} & U$_{\text{raster}}$, U$_{\text{raw}}$ $\times$ 3 & the interface effect on a canonical U-Net core \\
\bottomrule
\end{tabular}
\end{table}

\paragraph{Reporting priorities.}
Each study's commitment fixed its metric family, comparison pairs, and Holm
over the family and the bootstrap design, and none of that is changed here.
Every family is reported in full, with both estimands and both procedures, in
the supplement tables that the summary tables of Sec.~\ref{sec:results} cite.
In the text, each study is reported through its two primary endpoints,
calibrated selection NLL and calibrated completion log-loss, with the two
localisation metrics, probability mass within 5\,m and argmax distance,
alongside. Effect sizes and intervals come first and the max-$T$ adjusted $p$
is the prioritised inferential summary, subject to the assumptions above. Where a study has seeds, the per-seed standard
deviation of the pooled difference is printed beside the interval so that an
effect can be read against its own initialisation spread. This reporting choice was made after the studies had run; it relabels no
family and no procedure as pre-specified, and the Holm verdicts remain in
the supplement tables and their captions.

Studies 1, 2, 3, 4, 6, 8, 9, 11 and 13 and Part~B of Study 12 run on the
previously used eight-match holdout and are not evaluations on new data.
Studies 7 and 10 are the two studies on data from outside the tournament.
Study 5 first scores 56 of the 64 matches as test cases (Sec.~\ref{sec:threats}); Part~A of Study 12
scores the same 56 matches out of fold, against Study 5's locked
out-of-fold rows.

\section{Results}
\label{sec:results}

\begin{figure}[tbp]
\centering
\includegraphics[width=\textwidth]{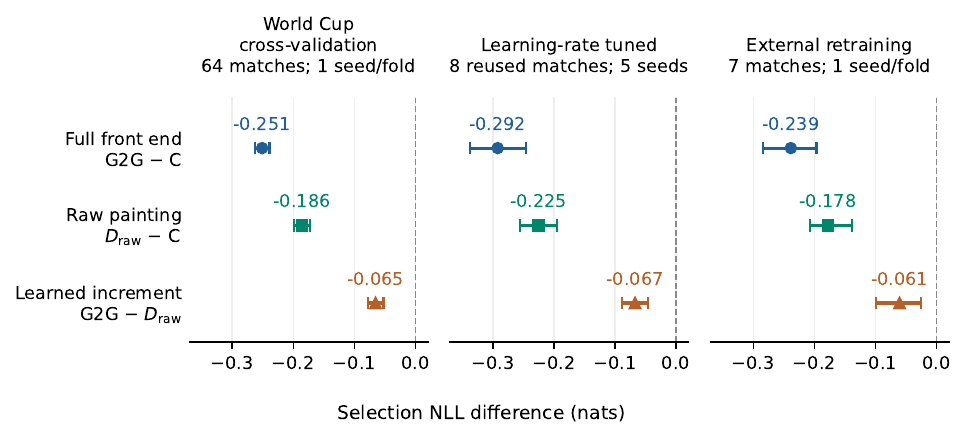}
\caption{Primary selection evidence. Points show pooled NLL differences
and bars show the reported pointwise 95\% bootstrap intervals; negative
values favour the first arm. The full front end is G2G minus C, raw painting
is D$_{\mathrm{raw}}$ minus C, and the learned increment is G2G minus
D$_{\mathrm{raw}}$. These panels reuse the results of Studies 5, 4, 7,
10, and 12, not new experiments. Cross-validation intervals condition on
one seed per fold; the tuned panel resamples matches and seeds; external
retraining has seven match clusters. Intervals are pointwise; the
family-adjusted tests are in the study tables}
\label{fig:primary}
\end{figure}

Secs.~\ref{sec:cv} and~\ref{sec:external} answer RQ1 of Sec.~\ref{sec:intro},
whether the painted front end improves on the same core across the tournament
and across a change of provider; Secs.~\ref{sec:tuned}--\ref{sec:setattn}
answer RQ2, which part of the front end carries the gain and whether it
persists under per-arm tuning. Estimands and metrics are defined in Secs.~\ref{sec:estimands}
and~\ref{sec:paired}; where the pooled and equal-match estimands differ, both
are shown. The primary comparisons report effect sizes and intervals; frozen-model
probes are descriptive sensitivity analyses. The max-$T$ sign-flip test is the prioritised inferential summary, exact
over all sign patterns in the seven- and eight-match studies and Monte Carlo
in the 64-match comparisons, and where a study has seeds the per-seed spread of
the effect is printed beside its interval (Sec.~\ref{sec:prespec}). Throughout, a contrast \emph{clears} a test, or \emph{survives} it, when its
adjusted $p$ is below $0.05$, and \emph{both procedures} means max-$T$ and
Holm. The subsections are ordered by evidential weight. The cross-validation over all 64 matches, the external replication and
the tuned, converged comparison (Secs.~\ref{sec:cv}, \ref{sec:external},
\ref{sec:tuned}) carry the paper's claim. The attribution ladder, the
front-end ablations, the raster controls and the interface alternatives
(Secs.~\ref{sec:frontend}--\ref{sec:setattn}) locate the mechanism on the reused holdout. The probes and
the tracking-noise curves (Sec.~\ref{sec:robustness}) test what the surfaces
rely on. The three-arm comparison from which the design came is reported as
development history (Sec.~\ref{sec:history}), and value, compute and the
surfaces themselves (Secs.~\ref{sec:value}--\ref{sec:surfaces}) close the
section. The pooled and equal-match estimands agree closely on every primary metric.
Figure~\ref{fig:primary} summarises the main selection contrasts;
Sec.~\ref{sec:compute} examines their computational trade-off.

\paragraph{Reading the summary tables.}
\label{par:reading}
Each main-text summary table gives the four headline metrics for each pair
shown of one study, in a common format; the full study tables, with every
pre-declared pair and metric, are in the supplement. Each pair spans two lines. The first gives the pooled
difference $\Delta$, the first arm minus the second, so a negative value
favours the first arm in the columns marked $\downarrow$ and a positive value
favours it in the column marked $\uparrow$, with its 95\% interval. The second
gives the per-seed standard deviation of that $\Delta$ where the study has
seeds, the number of test matches in which the first arm is ahead, and the
max-$T$ adjusted $p$, with $\circ$ for a max-$T$ rejection and $\ast$ for a
Holm rejection over the study's family at $\alpha=0.05$. Each caption then
gives only what is particular to its own study: the design, the bootstrap, the
size of the family, and the supplement table that prints the family in full.

\subsection{Cross-validation over the whole tournament}
\label{sec:cv}

The interface effect holds over the whole tournament: against the same core
fed rasters alone, \Gtwo{} improves selection NLL by $-0.251$ nats and leads
in every one of the 64 matches. Selection NLL is the mean negative log
probability the calibrated surface places on the cell the pass actually
reached, so a reduction of $d$ nats is a factor of $e^{d}$ in that probability on a
geometric average. The headline $0.251$ nats is therefore a
factor of about $1.29$: by that average \Gtwo{} places some $29\%$ more
probability on the observed destination than arm~C does. That is a relative statement
about one cell of a $7{,}072$-cell surface, not a claim about any downstream
decision. Every other World Cup study except Study 12 scores the same eight matches
(Sec.~\ref{sec:history}); Study 5 (Sec.~\ref{sec:prespec})
scores all 64 in eight match-grouped folds, the shipped test set as fold 1
and the other 56 matches assigned by the same label-blind hash rule, each
fold's four arms trained at the shipped recipe with one seed and tested on
the fold they never saw. Arm~C and \Gtwo{} differ only in the painted front
end, so their contrast is the interface effect. Table~\ref{tab:summary-cv}
summarises the family; Supplement Tables~\suppref{tab:cv}
and~\suppref{tab:cv-folds} give all 50 hypotheses and the two primary
metrics fold by fold.

\paragraph{The effect holds on the 56 matches never before scored as test data.}
Pooled over the tournament, \Gtwo{} reaches a selection NLL of $5.024$
against $5.275$ for arm~C, and its margin over the two SoccerMap controls is
between half a nat and seven tenths; completion log-loss improves by one to
three hundredths against every raster arm. \Gtwo{} wins all 64 matches on
selection NLL against every raster arm and on every headline metric against
both SoccerMap controls. With 64 clusters the smallest adjusted $p$ this test can return is
$10^{-5}$ rather than the $2/256=0.0078$ of the eight-match studies, and every
headline contrast of the three \Gtwo{} pairs clears both procedures at that
floor; only the five calibration-error contrasts do not. Tables~\ref{tab:null} and~\ref{tab:null-completion} give these levels a
scale, one for each supervised head, by scoring two models that need no
network on the same rows and folds. On selection, against a uniform surface
and against a displacement prior fitted on each fold's training matches,
every learned arm is ahead by more than two nats in all 64 matches, so the
interface effect is about a tenth of the distance between the raster core and
the stronger null. On completion the same construction puts the interface effect at about a
fourteenth of the distance to its own null: the length prior reaches $0.397$
of log-loss where arm~C reaches $0.189$ and \Gtwo{} $0.174$, so the $-0.015$
the front end adds sits against a $-0.207$ gap between the prior and the
core. Both nulls are deliberately simple, and a richer network-free baseline
would narrow these ratios. The controls order differently on the two heads:
on selection the wide control is ahead of the original, while on completion
the two are level, and both trail arm~C.

\begin{table}[htb]
\caption{Selection null models under the Study 5 protocol: pooled out-of-fold
metrics over all 53{,}628 passes and 64 matches. \emph{Uniform} spreads its
mass over the 7{,}072 cells. The \emph{displacement prior} is a smoothed
two-dimensional histogram of the endpoint's displacement from the ball's cell
at release, estimated on each fold's 48 training matches with the smoothing
width chosen on its 8 validation matches, so it uses no test row and no
network. Arm rows are the locked Study 5 runs of Table~\ref{tab:summary-cv}.
Every learned arm is ahead of both nulls in all 64 matches; the gap of arm~C
over the displacement prior is $-2.601$ nats $[-2.659,-2.542]$ (64-cluster
match bootstrap, $B=10{,}000$) and of \Gtwo{} $-2.852$ $[-2.913,-2.791]$}
\label{tab:null}
\centering\small
\begin{tabular}{lrrr}
\toprule
Model & Selection NLL & Mass $<$5\,m & Argmax dist.\ (m) \\
\midrule
Uniform over the grid & 8.864 & --- & --- \\
Displacement prior & 7.876 & 0.038 & 18.1 \\
Original SoccerMap & 5.727 & 0.214 & 14.0 \\
Wide SoccerMap & 5.544 & 0.233 & 13.4 \\
Arm C (\vtwo{}) & 5.275 & 0.276 & 12.1 \\
\Gtwo{} & 5.024 & 0.304 & 11.3 \\
\bottomrule
\end{tabular}
\end{table}

\begin{table}[htb]
\caption{Completion null models under the Study 5 protocol: pooled
out-of-fold metrics over the same 53{,}628 passes and 64 matches as
Table~\ref{tab:null}, which gives the selection counterpart. The
\emph{base rate} is one constant, the completion rate of each fold's 48
training matches. The \emph{length prior} is completion probability as a
function of the straight release-to-endpoint distance, an empirical curve
over 1\,m bins estimated on those same training matches with the
smoothing width chosen on the fold's 8 validation matches, so it uses no
test row and no network. Arm rows are the locked Study 5 runs of
Table~\ref{tab:summary-cv}, temperature-scaled on validation. The base
rate is one constant within each fold, so its AUC carries no ranking
information and is omitted. Every learned arm is ahead of both nulls in
all 64 matches; the gap of arm~C over the length prior is
${-}0.207$ $[{-}0.216,{-}0.199]$ nats and of \Gtwo{} ${-}0.223$ $[{-}0.231,{-}0.215]$ (64-cluster match bootstrap, $B=10{,}000$)}
\label{tab:null-completion}
\centering\small
\begin{tabular}{lrrr}
\toprule
Model & Completion log-loss & Brier & AUC \\
\midrule
Base rate (one constant) & 0.444 & 0.136 & --- \\
Length prior & 0.397 & 0.119 & 0.685 \\
Original SoccerMap & 0.206 & 0.061 & 0.945 \\
Wide SoccerMap & 0.208 & 0.061 & 0.944 \\
Arm C (\vtwo{}) & 0.189 & 0.056 & 0.955 \\
\Gtwo{} & 0.174 & 0.051 & 0.962 \\
\bottomrule
\end{tabular}
\end{table}

\paragraph{Fold 1 is the hardest fold for the painted arms.}
Fold 1, the eight matches the holdout studies reuse, is the hardest fold on
selection for \Gtwo{} and arm~C and among the harder ones for the two
SoccerMap controls, whose worst folds are 2 and 7, and \Gtwo{}'s margin over
the original control there is below the tournament-wide one (Supplement
Table~\suppref{tab:cv-folds}). Two qualifications follow. The learning rates, budget and architecture were
chosen on the original split, so folds 2--8 are unseen as test data but not
as data the design was tuned against. A fold's validation matches serve as
the next fold's test matches, the standard rotation, which leaks no label
but means the eight models are not independent.

\paragraph{The core and the front end again.}
Across the tournament the modernised core alone improves on both controls
on selection and on completion under both procedures, and the front end
adds its quarter of a nat on top; the two steps are the same size here as
in the single-split studies, as one expects if they are properties of the
architectures rather than of the eight matches.

\paragraph{The recommended variant under the same protocol.}
Study 12, Part A (Sec.~\ref{sec:prespec}) puts the raw-feature painter
D$_{\text{raw}}$ through the same eight folds at the same recipe and scores
it against the locked out-of-fold rows of C and \Gtwo{}
(Table~\ref{tab:summary-cv-raw}; Supplement Tables~\suppref{tab:cv-raw}
and~\suppref{tab:cv-raw-folds}). With no learned front-end parameter at
all, painting the nine raw node features is worth $-0.186$ nats of
selection NLL $[-0.199,-0.172]$ over arm~C, ahead in all 64 matches and
under both procedures at the same $10^{-5}$ floor, three quarters of the $-0.251$ the full front end
earns, the same share it carries on the external corpus
(Sec.~\ref{sec:external}). The remaining quarter, $-0.065$
$[-0.078,-0.052]$, is what the learned encoder and message passing add
together; over 64 match clusters that increment clears both procedures on every
metric of the family except calibration error, and it is resolved again under
per-arm tuning with five seeds
(Sec.~\ref{sec:tuned}), whereas in the three-seed holdout studies it sits
within seed noise (Secs.~\ref{sec:attribution}, \ref{sec:ablation}). The
tournament-wide reading is therefore sharper than the holdout's: the write
carries three quarters of the gain, the learned content one quarter.

\begin{table}[t]
\caption{Study 5 summary: the eight-fold match-grouped cross-validation over all 64 matches at the shipped recipe, pooled out-of-fold predictions of 53{,}628 passes in 64 match clusters, one seed per fold. Intervals from the 64-cluster match bootstrap over 64 matches ($B=10{,}000$); family of 50 hypotheses. Reading conventions: Sec.~\ref{par:reading}. Full family: Supplement Table~\suppref{tab:cv}}
\label{tab:summary-cv}
\centering
\footnotesize
\setlength{\tabcolsep}{2pt}
\renewcommand{\arraystretch}{1.0}
\begin{tabular}{lrrrr}
\toprule
Pair & Selection NLL\,$\downarrow$ & Completion log-loss\,$\downarrow$ & Mass $<$5\,m\,$\uparrow$ & Argmax dist.\ (m)\,$\downarrow$ \\
\midrule
\Gtwo{}$-$C & $-0.251$ $[-0.263,-0.239]$ & $-0.015$ $[-0.018,-0.013]$ & $+0.028$ $[+0.026,+0.031]$ & $-0.77$ $[-0.85,-0.68]$ \\
{\scriptsize\itshape matches, adj.\ $p$} & {\scriptsize 64/64, $p<0.01$ $\ast$$\circ$} & {\scriptsize 58/64, $p<0.01$ $\ast$$\circ$} & {\scriptsize 64/64, $p<0.01$ $\ast$$\circ$} & {\scriptsize 62/64, $p<0.01$ $\ast$$\circ$} \\[1pt]
\Gtwo{}$-$original & $-0.704$ $[-0.736,-0.670]$ & $-0.032$ $[-0.035,-0.029]$ & $+0.090$ $[+0.084,+0.097]$ & $-2.74$ $[-2.90,-2.58]$ \\
{\scriptsize\itshape matches, adj.\ $p$} & {\scriptsize 64/64, $p<0.01$ $\ast$$\circ$} & {\scriptsize 64/64, $p<0.01$ $\ast$$\circ$} & {\scriptsize 64/64, $p<0.01$ $\ast$$\circ$} & {\scriptsize 64/64, $p<0.01$ $\ast$$\circ$} \\[1pt]
\Gtwo{}$-$wide & $-0.520$ $[-0.553,-0.487]$ & $-0.034$ $[-0.037,-0.031]$ & $+0.071$ $[+0.065,+0.078]$ & $-2.15$ $[-2.30,-2.00]$ \\
{\scriptsize\itshape matches, adj.\ $p$} & {\scriptsize 64/64, $p<0.01$ $\ast$$\circ$} & {\scriptsize 64/64, $p<0.01$ $\ast$$\circ$} & {\scriptsize 64/64, $p<0.01$ $\ast$$\circ$} & {\scriptsize 64/64, $p<0.01$ $\ast$$\circ$} \\[1pt]
C$-$original & $-0.452$ $[-0.480,-0.425]$ & $-0.017$ $[-0.019,-0.014]$ & $+0.062$ $[+0.057,+0.067]$ & $-1.97$ $[-2.11,-1.84]$ \\
{\scriptsize\itshape matches, adj.\ $p$} & {\scriptsize 64/64, $p<0.01$ $\ast$$\circ$} & {\scriptsize 61/64, $p<0.01$ $\ast$$\circ$} & {\scriptsize 64/64, $p<0.01$ $\ast$$\circ$} & {\scriptsize 64/64, $p<0.01$ $\ast$$\circ$} \\[1pt]
C$-$wide & $-0.269$ $[-0.296,-0.241]$ & $-0.019$ $[-0.022,-0.016]$ & $+0.043$ $[+0.038,+0.048]$ & $-1.38$ $[-1.51,-1.24]$ \\
{\scriptsize\itshape matches, adj.\ $p$} & {\scriptsize 63/64, $p<0.01$ $\ast$$\circ$} & {\scriptsize 62/64, $p<0.01$ $\ast$$\circ$} & {\scriptsize 63/64, $p<0.01$ $\ast$$\circ$} & {\scriptsize 64/64, $p<0.01$ $\ast$$\circ$} \\
\bottomrule
\end{tabular}
\end{table}

\begin{table}[t]
\caption{Study 12 (A) summary: the raw-feature arm D$_{\text{raw}}$ in the Study 5 eight-fold match-grouped cross-validation at the shipped recipe, one seed per fold, against the locked Study 5 out-of-fold rows of C and \Gtwo{}. Intervals from the 64-cluster match bootstrap over 64 matches ($B=10{,}000$); family of 20 hypotheses. Reading conventions: Sec.~\ref{par:reading}. Full family: Supplement Table~\suppref{tab:cv-raw}}
\label{tab:summary-cv-raw}
\centering
\footnotesize
\setlength{\tabcolsep}{2pt}
\renewcommand{\arraystretch}{1.0}
\begin{tabular}{lrrrr}
\toprule
Pair & Selection NLL\,$\downarrow$ & Completion log-loss\,$\downarrow$ & Mass $<$5\,m\,$\uparrow$ & Argmax dist.\ (m)\,$\downarrow$ \\
\midrule
D$_{\text{raw}}$$-$C & $-0.186$ $[-0.199,-0.172]$ & $-0.010$ $[-0.012,-0.007]$ & $+0.017$ $[+0.015,+0.020]$ & $-0.46$ $[-0.55,-0.37]$ \\
{\scriptsize\itshape matches, adj.\ $p$} & {\scriptsize 64/64, $p<0.01$ $\ast$$\circ$} & {\scriptsize 53/64, $p<0.01$ $\ast$$\circ$} & {\scriptsize 63/64, $p<0.01$ $\ast$$\circ$} & {\scriptsize 57/64, $p<0.01$ $\ast$$\circ$} \\[1pt]
\Gtwo{}$-$D$_{\text{raw}}$ & $-0.065$ $[-0.078,-0.052]$ & $-0.006$ $[-0.008,-0.003]$ & $+0.011$ $[+0.008,+0.014]$ & $-0.31$ $[-0.41,-0.21]$ \\
{\scriptsize\itshape matches, adj.\ $p$} & {\scriptsize 55/64, $p<0.01$ $\ast$$\circ$} & {\scriptsize 43/64, $p<0.01$ $\ast$$\circ$} & {\scriptsize 56/64, $p<0.01$ $\ast$$\circ$} & {\scriptsize 53/64, $p<0.01$ $\ast$$\circ$} \\
\bottomrule
\end{tabular}
\end{table}

\subsection{External replication on public German league tracking data}
\label{sec:external}

Retrained on seven public Bundesliga and 2.~Bundesliga matches from
another provider, league and season, \Gtwo{} leads the same core by $-0.239$ nats of selection NLL in every
match; frozen, it keeps its localisation advantage but not its selection
likelihood. Every study but this one and Study 10 draws on one tournament from one
provider. Study 7 (Sec.~\ref{sec:prespec}) takes the frozen pipeline to a
public corpus from another league, tracking provider and season and asks
whether the locked World Cup models keep their ordering when scored,
unchanged, on passes they were never trained for, and whether the ordering
reappears when every arm is trained from scratch on the external matches
alone.

\paragraph{The external benchmark.}
The IDSSE release~\citep{bassek2025idsse} holds seven complete Bundesliga
and 2.~Bundesliga matches of season 2022/23 with raw event data and raw
optical tracking at 25\,Hz, under CC~BY~4.0; no model of this paper had seen
a frame of it before Study 7. We imported the World Cup feature chain
verbatim and re-implemented the admission gates on the IDSSE schema, with an
estimated per-match event--frame alignment and a receiver-capture-first
endpoint rule; the port admits 4{,}241 passes, and PE is not evaluated, no
value target having been ported. Supplement Sec.~\suppref{supp:external}
gives the port in full, with its gate audit and every deviation from the
World Cup pipeline; the corpus is also described, as a benchmark, in a companion manuscript by the authors
(Sec.~\ref{sec:intro}), none of whose results this paper uses. Its endpoints
are reconstructed from tracking under an estimated alignment rather than
annotated, so what follows is a change-of-provider replication on a
reconstructed corpus, not a measurement against clean labels.

\paragraph{Design.}
Part~A is zero-shot: the locked World Cup checkpoints of \Gtwo{}, arm~C and
both SoccerMap controls, three seeds each, are scored with their locked
temperatures on every external pass, nothing refitted. Part~B retrains all
four arms from scratch at the shipped recipe in seven leave-one-match-out
folds, one seed per fold plus a three-seed repeat of fold 1 that enters no
family, every run hashed before any test match is opened. Each part is a
family of 30 hypotheses, \Gtwo{} against original, wide and C on the ten PP
and PS metrics, with a hierarchical bootstrap over the 7 matches (and, in
Part~A, the 3 seeds) and the prioritised sign-flip test over all $2^{7}=128$
match-level sign patterns, whose smallest attainable adjusted $p$ is
$2/128=0.0156$. Table~\ref{tab:summary-external} summarises both parts;
Supplement Tables~\suppref{tab:external-a}--\suppref{tab:external-fold1}
give the families, the primary metrics fold by fold and the fold-1 repeat.

\begin{table}[t]
\caption{Study 7 summary: the external replication on the seven IDSSE matches (4{,}241 passes, 7 match clusters). Part A scores the locked World Cup checkpoints (three seeds per arm) with their locked temperatures, nothing refitted; Part B retrains every arm from scratch in seven leave-one-match-out folds, one seed per fold. Intervals from the hierarchical bootstrap over 7 matches and 3 seeds (Part A) or 7-cluster match bootstrap (Part B) ($B=10{,}000$); family of 30 hypotheses per part; $\dagger$: SD from the three-seed fold-1 repeat. Reading conventions: Sec.~\ref{par:reading}. Full family: Supplement Tables~\suppref{tab:external-a} and~\suppref{tab:external-b}}
\label{tab:summary-external}
\centering
\footnotesize
\setlength{\tabcolsep}{2pt}
\renewcommand{\arraystretch}{1.0}
\begin{tabular}{lrrrr}
\toprule
Pair & Selection NLL\,$\downarrow$ & Completion log-loss\,$\downarrow$ & Mass $<$5\,m\,$\uparrow$ & Argmax dist.\ (m)\,$\downarrow$ \\
\midrule
\multicolumn{5}{l}{\emph{zero-shot (Part A)}} \\
\Gtwo{}$-$C & $+0.079$ $[+0.036,+0.121]$ & $+0.011$ $[-0.003,+0.028]$ & $+0.007$ $[+0.001,+0.014]$ & $-0.01$ $[-0.21,+0.18]$ \\
{\scriptsize\itshape matches, adj.\ $p$} & {\scriptsize sd 0.042, 0/7, $p=0.016$ $\ast$$\circ$} & {\scriptsize sd 0.013, 1/7, $p=0.469$} & {\scriptsize sd 0.006, 6/7, $p=0.203$} & {\scriptsize sd 0.06, 4/7, $p=1$} \\[1pt]
\Gtwo{}$-$original & $-0.045$ $[-0.144,+0.039]$ & $-0.003$ $[-0.014,+0.010]$ & $+0.044$ $[+0.035,+0.052]$ & $-0.68$ $[-1.20,-0.28]$ \\
{\scriptsize\itshape matches, adj.\ $p$} & {\scriptsize sd 0.085, 4/7, $p=0.797$} & {\scriptsize sd 0.005, 4/7, $p=1$} & {\scriptsize sd 0.006, 7/7, $p=0.016$ $\ast$$\circ$} & {\scriptsize sd 0.27, 7/7, $p=0.219$ $\ast$} \\[1pt]
\Gtwo{}$-$wide & $+0.033$ $[-0.018,+0.084]$ & $+0.010$ $[-0.005,+0.025]$ & $+0.031$ $[+0.024,+0.037]$ & $-0.64$ $[-1.06,-0.28]$ \\
{\scriptsize\itshape matches, adj.\ $p$} & {\scriptsize sd 0.037, 2/7, $p=0.969$} & {\scriptsize sd 0.006, 1/7, $p=0.922$} & {\scriptsize sd 0.003, 7/7, $p=0.016$ $\ast$$\circ$} & {\scriptsize sd 0.21, 6/7, $p=0.203$ $\ast$} \\
\midrule
\multicolumn{5}{l}{\emph{retrained (Part B)}} \\
\Gtwo{}$-$C & $-0.239$ $[-0.285,-0.197]$ & $-0.026$ $[-0.030,-0.022]$ & $+0.044$ $[+0.035,+0.055]$ & $-0.83$ $[-1.15,-0.51]$ \\
{\scriptsize\itshape matches, adj.\ $p$} & {\scriptsize sd 0.032$^{\dagger}$, 7/7, $p=0.016$ $\ast$$\circ$} & {\scriptsize sd 0.002$^{\dagger}$, 7/7, $p=0.016$ $\ast$$\circ$} & {\scriptsize sd 0.002$^{\dagger}$, 7/7, $p=0.031$ $\ast$$\circ$} & {\scriptsize sd 0.28$^{\dagger}$, 7/7, $p=0.047$ $\ast$$\circ$} \\[1pt]
\Gtwo{}$-$original & $-0.887$ $[-1.024,-0.742]$ & $-0.041$ $[-0.064,-0.018]$ & $+0.136$ $[+0.116,+0.153]$ & $-2.59$ $[-2.88,-2.27]$ \\
{\scriptsize\itshape matches, adj.\ $p$} & {\scriptsize sd 0.050$^{\dagger}$, 7/7, $p=0.016$ $\ast$$\circ$} & {\scriptsize sd 0.038$^{\dagger}$, 7/7, $p=0.094$ $\ast$} & {\scriptsize sd 0.009$^{\dagger}$, 7/7, $p=0.016$ $\ast$$\circ$} & {\scriptsize sd 0.08$^{\dagger}$, 7/7, $p=0.016$ $\ast$$\circ$} \\[1pt]
\Gtwo{}$-$wide & $-0.591$ $[-0.678,-0.508]$ & $-0.027$ $[-0.041,-0.015]$ & $+0.100$ $[+0.083,+0.117]$ & $-2.83$ $[-3.57,-2.11]$ \\
{\scriptsize\itshape matches, adj.\ $p$} & {\scriptsize sd 0.315$^{\dagger}$, 7/7, $p=0.016$ $\ast$$\circ$} & {\scriptsize sd 0.004$^{\dagger}$, 6/7, $p=0.062$ $\ast$} & {\scriptsize sd 0.034$^{\dagger}$, 7/7, $p=0.016$ $\ast$$\circ$} & {\scriptsize sd 0.92$^{\dagger}$, 7/7, $p=0.031$ $\ast$$\circ$} \\
\bottomrule
\end{tabular}
\end{table}

\paragraph{Zero-shot transfer.}
Frozen, the World Cup models find the external corpus harder by more than a
nat of selection NLL each, and the shift does not treat the arms alike
(Table~\ref{tab:summary-external}, Part~A). On selection NLL \Gtwo{} is level
with both SoccerMap controls and behind arm~C in every one of the seven
matches, a contrast that clears the max-$T$ test in C's favour;
completion is a wash for all three pairs. What does transfer is where the
mass goes: \Gtwo{} puts more probability within 5\,m of the observed
endpoint than either SoccerMap control, ahead in all seven matches and
clearing the max-$T$ test, and its argmax lands closer. The frozen
models thus keep the sharper surfaces the splat produces but not the better
selection likelihood, on the raw score as well as the calibrated one
(Supplement Table~\suppref{tab:external-a}); the raster-only core degrades
most gracefully.

\paragraph{Retrained on five matches.}
Fitted to the new distribution, the full ordering returns
(Table~\ref{tab:summary-external}, Part~B). Out of fold, \Gtwo{} scores
$6.377$ nats of selection NLL against $6.616$ for arm~C, and it leads all
three raster arms in all seven matches on selection NLL and local mass under
both procedures, by more than half a nat against the SoccerMap controls.
Completion follows by a few hundredths of log-loss against every arm, but at
the $0.0156$ floor of the seven-cluster test only the contrast against C
clears the adjusted max-$T$ threshold; the calibration-error contrasts clear neither
procedure. \Gtwo{} is ahead of arm~C on both primary metrics in every fold
(Supplement Table~\suppref{tab:external-folds}), and the fold-1 repeat
(Supplement Table~\suppref{tab:external-fold1}) puts the seed noise of the
\Gtwo{}$-$C contrasts well inside their margins, all three seeds in favour;
the contrast against the wide control is the variable one, the wide
control's own convergence on about three thousand training passes being the
variable part.

\paragraph{What transfers.}
The front-end margin over the same core, \Gtwo{}$-$C, is $-0.239$ nats on the
external corpus and $-0.251$ on the 64-match cross-validation of
Sec.~\ref{sec:cv}: trained on about three thousand passes from another
provider, league and season, with an estimated event--frame alignment and raw
optical positions, the entity-to-grid interface is worth what it was worth on
the World Cup. The margin over the SoccerMap controls, core and front end
together, is larger externally; the controls are the arms that suffer most
from five training matches.

\paragraph{The raw-feature arm on the external corpus.}
Study 7 carries the shipped model and three controls across the provider
boundary but not the front-end ablation that matters most for the interface
claim: D$_{\text{raw}}$, the nine raw node features painted with no learned
encoder (Sec.~\ref{sec:ablation}), which on the World Cup sits within a
hundredth of a nat of the learned embeddings. Study 10
(Sec.~\ref{sec:prespec}) adds that arm to both parts under the Study 7
protocol with nothing else changed, its three locked Study 3 checkpoints
scored zero-shot against the locked Study 7 rows of C and \Gtwo{} (a family
of 20) and its seven retrained folds scored against the locked rows of C,
\Gtwo{} and original (a family of 30), with the intervals, test and floor of
Study 7. Table~\ref{tab:summary-external-raw} summarises both parts;
Supplement
Tables~\suppref{tab:external-raw-a}--\suppref{tab:external-raw-fold1} give
the families, the fold-by-fold metrics and the fold-1 repeat.

\begin{table}[t]
\caption{Study 10 summary: the raw-feature arm D$_{\text{raw}}$ on the seven IDSSE matches under the Study 7 protocol, against the locked Study 7 rows of C, \Gtwo{} and original: Part A zero-shot with the three locked Study 3 checkpoints, Part B retrained in the seven leave-one-match-out folds. Intervals from the hierarchical bootstrap over 7 matches and 3 seeds (Part A) or 7-cluster match bootstrap (Part B) ($B=10{,}000$); family of 20 (Part A) and 30 (Part B) hypotheses per part; $\dagger$: SD from the three-seed fold-1 repeat. Reading conventions: Sec.~\ref{par:reading}. Full family: Supplement Tables~\suppref{tab:external-raw-a} and~\suppref{tab:external-raw-b}}
\label{tab:summary-external-raw}
\centering
\footnotesize
\setlength{\tabcolsep}{2pt}
\renewcommand{\arraystretch}{1.0}
\begin{tabular}{lrrrr}
\toprule
Pair & Selection NLL\,$\downarrow$ & Completion log-loss\,$\downarrow$ & Mass $<$5\,m\,$\uparrow$ & Argmax dist.\ (m)\,$\downarrow$ \\
\midrule
\multicolumn{5}{l}{\emph{zero-shot (Part A)}} \\
D$_{\text{raw}}$$-$C & $+0.030$ $[-0.021,+0.084]$ & $-0.007$ $[-0.017,+0.003]$ & $+0.006$ $[+0.000,+0.010]$ & $-0.14$ $[-0.32,+0.03]$ \\
{\scriptsize\itshape matches, adj.\ $p$} & {\scriptsize sd 0.050, 1/7, $p=0.281$} & {\scriptsize sd 0.009, 7/7, $p=0.016$ $\circ$} & {\scriptsize sd 0.003, 6/7, $p=0.328$} & {\scriptsize sd 0.03, 5/7, $p=0.453$} \\[1pt]
\Gtwo{}$-$D$_{\text{raw}}$ & $+0.048$ $[+0.023,+0.070]$ & $+0.018$ $[+0.006,+0.030]$ & $+0.002$ $[-0.006,+0.009]$ & $+0.14$ $[-0.03,+0.35]$ \\
{\scriptsize\itshape matches, adj.\ $p$} & {\scriptsize sd 0.012, 0/7, $p=0.062$ $\ast$} & {\scriptsize sd 0.009, 0/7, $p=0.141$} & {\scriptsize sd 0.008, 5/7, $p=0.969$} & {\scriptsize sd 0.05, 1/7, $p=0.641$} \\
\midrule
\multicolumn{5}{l}{\emph{retrained (Part B)}} \\
D$_{\text{raw}}$$-$C & $-0.178$ $[-0.207,-0.139]$ & $-0.018$ $[-0.023,-0.013]$ & $+0.030$ $[+0.021,+0.038]$ & $-0.62$ $[-0.80,-0.41]$ \\
{\scriptsize\itshape matches, adj.\ $p$} & {\scriptsize sd 0.077$^{\dagger}$, 7/7, $p=0.016$ $\ast$$\circ$} & {\scriptsize sd 0.002$^{\dagger}$, 7/7, $p=0.016$ $\ast$$\circ$} & {\scriptsize sd 0.013$^{\dagger}$, 7/7, $p=0.016$ $\ast$$\circ$} & {\scriptsize sd 0.45$^{\dagger}$, 7/7, $p=0.016$ $\ast$$\circ$} \\[1pt]
\Gtwo{}$-$D$_{\text{raw}}$ & $-0.061$ $[-0.099,-0.025]$ & $-0.008$ $[-0.013,-0.004]$ & $+0.014$ $[+0.006,+0.023]$ & $-0.20$ $[-0.57,+0.12]$ \\
{\scriptsize\itshape matches, adj.\ $p$} & {\scriptsize sd 0.081$^{\dagger}$, 6/7, $p=0.109$ $\ast$} & {\scriptsize sd 0.003$^{\dagger}$, 6/7, $p=0.172$ $\ast$} & {\scriptsize sd 0.011$^{\dagger}$, 6/7, $p=0.156$ $\ast$} & {\scriptsize sd 0.28$^{\dagger}$, 4/7, $p=0.859$} \\[1pt]
D$_{\text{raw}}$$-$original & $-0.826$ $[-0.970,-0.673]$ & $-0.033$ $[-0.059,-0.006]$ & $+0.122$ $[+0.100,+0.144]$ & $-2.39$ $[-2.53,-2.26]$ \\
{\scriptsize\itshape matches, adj.\ $p$} & {\scriptsize sd 0.042$^{\dagger}$, 7/7, $p=0.016$ $\ast$$\circ$} & {\scriptsize sd 0.038$^{\dagger}$, 5/7, $p=0.453$} & {\scriptsize sd 0.008$^{\dagger}$, 7/7, $p=0.016$ $\ast$$\circ$} & {\scriptsize sd 0.29$^{\dagger}$, 7/7, $p=0.016$ $\ast$$\circ$} \\
\bottomrule
\end{tabular}
\end{table}

\paragraph{Zero-shot transfer of the raw-feature arm.}
Frozen, the raw-feature arm has better likelihood point estimates than G2G
(Table~\ref{tab:summary-external-raw}, Part~A). Its selection NLL sits between arm~C's and \Gtwo{}'s: against C it is a null
contrast, whereas \Gtwo{} is behind it in all seven matches, without an
adjusted max-$T$ rejection ($p=0.062$). On completion the frozen raw-feature
arm is ahead of C in all seven matches and clears the max-$T$ test, while its
margin over \Gtwo{} does not, and local mass is level with \Gtwo{}'s. The frozen raw-feature painter thus keeps the splat's sharper
surfaces across the provider boundary as \Gtwo{} does, but its calibrated
likelihood degrades as arm~C's does rather than as \Gtwo{}'s, a pattern
consistent with a learned encoder fitted to the source distribution.

\paragraph{Retrained on five matches without a learned encoder.}
Retrained on five matches, the raw-feature arm reproduces the World Cup
attribution (Table~\ref{tab:summary-external-raw}, Part~B). Out of fold it
scores $6.438$ nats of selection NLL against $6.616$ for arm~C and $6.377$
for \Gtwo{}: it leads C and the original control in all seven matches on
selection NLL, local mass and argmax distance under both procedures, and
\Gtwo{} keeps a small margin above it on selection, completion and local
mass, six of seven matches each, short of the adjusted max-$T$ threshold on all three.
D$_{\text{raw}}$ sits between C and \Gtwo{} on selection in six of the seven
folds (Supplement Table~\suppref{tab:external-raw-folds}), and in the fold-1
repeat (Supplement Table~\suppref{tab:external-raw-fold1}) all three seeds
favour D$_{\text{raw}}$ over C and \Gtwo{} over D$_{\text{raw}}$, the second
increment being of the order of its own seed noise. Of the $0.239$ nats that
separate \Gtwo{} from arm~C on the external corpus, the raw-feature painter
with no learned parameters accounts for about three quarters, the share the
painted package holds on the World Cup (Sec.~\ref{sec:attribution}). On both corpora raw painting accounts for most of the measured gain;
the additional learned-content increment is resolved in the World Cup
cross-validation and tuned studies, but not in this small external comparison.

\paragraph{The external effect and the endpoint rule.}
The two corpora resolve their endpoints differently: on the World Cup
$98.9\%$ of admitted passes end at the next tracked possession event and
$0.21\%$ at a receiver capture, whereas on the external corpus the mix is
inverted, $91.2\%$ capture and $6.6\%$ next event (Supplement
Table~\suppref{tab:external-data}). Because the capture rule returns the
first frame at which the ball is within $1$\,m of the annotated receiver,
an external endpoint lies within about one grid cell of a tracked player by
construction, which is a label a front end that writes player features at
continuous coordinates might exploit. We therefore re-scored the locked
Part~B out-of-fold rows inside each stratum, a post-hoc analysis that no
commitment record covers. The effect is carried by the capture stratum:
there \Gtwo{} leads arm~C by $-0.266$ nats $[-0.323,-0.216]$ in all seven
matches, while on the $282$ next-event rows the point
estimate turns round to $+0.102$, with an interval $[-0.013,+0.247]$ that
spans zero and two of seven matches favouring \Gtwo{}, and on the $91$
boundary rows it is $-0.128$ $[-0.452,+0.278]$. These are pointwise intervals over the same seven match clusters, with no
sign-flip test and no adjustment across the strata. The strata are also not
a clean contrast: the next-event rows are the harder ones, arm~C scoring
$7.60$ nats on them against $6.50$ in the capture stratum, they are few, and
they differ from the capture rows in completion rate and in pass length, so
endpoint provenance is confounded with what kind of pass it is. Two readings
survive: the external margin may be partly a property of the capture rule,
and the World Cup result, whose labels are almost entirely next-event, shows
that the interface effect does not require that rule. What the external study
cannot do is separate the two.

\subsection{Tuned, converged, five-seed comparison}
\label{sec:tuned}

With every arm tuned to its own learning rate and trained to convergence,
\Gtwo{} stays $-0.292$ nats of selection NLL ahead of the same core, and the
margin against both tuned SoccerMap controls is about $0.3$ nats. The shipped
comparison trains every arm at one shared learning rate under one epoch cap,
which Supplement Sec.~\suppref{sec:lrsweep} shows favours \Gtwo{} on
validation. Study 4 (Sec.~\ref{sec:prespec}) removes both conditions at
once: we trained each arm's PP and PS heads at their own validation-selected
learning rate to a cap of 200 epochs under five seeds, and opened the test
set once after every checkpoint was locked. Table~\ref{tab:summary-tuned}
summarises the family; Supplement Table~\suppref{tab:tuned} gives all 50
hypotheses under both estimands and both procedures.

\paragraph{Selected rates and convergence.}
The committed rule chose $2\times10^{-4}$ for \Gtwo{}'s selection head,
$10^{-4}$ for arm~C's and the wide control's and $5\times10^{-5}$ for the
original control's, and completion-head rates between $10^{-4}$ and
$5\times10^{-4}$; the shipped completion rate was the bottom of its grid and
the shipped selection rate lies below it. Every evaluated head run stopped
early inside the cap. Thus the cap is not binding under the selected rates
and stopping rule; this does not establish an optimiser-independent optimum.

\paragraph{The gap narrows and stays.}
Seed-averaged over five seeds, \Gtwo{} reaches a selection NLL of $4.965$
against $5.257$ for tuned arm~C, with both tuned SoccerMap controls within a
few hundredths of arm~C (Table~\ref{tab:summary-tuned}); every selection and
completion contrast of the three \Gtwo{} pairs clears both procedures, the
calibration-error contrasts excepted. Per-arm tuning therefore takes the
matched-budget selection margin against the capacity-matched control from
about half a nat (Supplement Table~\suppref{tab:paired-full}) to a third,
where the cap-200 reruns of Supplement Sec.~\suppref{sec:convergence} put
it, and leaves the completion margin at three-quarters of its shipped size:
what tuning removes is the part of the margin the shared recipe had given
\Gtwo{} for free, and what it leaves is the part the paper claims.

\paragraph{Arm C under tuning.}
Tuned, the modernised core beats both tuned controls on completion log-loss
and argmax distance under both procedures, but not on selection NLL, where
both of its contrasts span zero. Under tuning the core alone is therefore a
better completion model than SoccerMap but not a better selection model; it
is the painted front end that separates \Gtwo{} from all three raster arms
on selection by about $0.3$ nats.

\paragraph{The recommended variant, tuned.}
Study 12, Part B (Sec.~\ref{sec:prespec}) runs D$_{\text{raw}}$ through the
same sweep, selection rule and five seeds against the locked tuned runs of
C and \Gtwo{} (Table~\ref{tab:summary-tuned-raw}; Supplement
Table~\suppref{tab:tuned-raw}). The committed rule chose $10^{-4}$ for its
completion head and $2\times10^{-4}$ for its selection head. Tuned and
converged, the raw-feature painter is $-0.225$ nats $[-0.256,-0.195]$
ahead of the tuned core on selection NLL, in all eight matches under both
procedures, three quarters of the tuned front-end margin; \Gtwo{} keeps
$-0.067$ $[-0.088,-0.045]$ over it, again in all eight matches under both
procedures, with completion level between the two. Under tuning, then, the
split is the one the cross-validation gives: the write carries three
quarters of the gain and the learned content a quarter.

\begin{table}[t]
\caption{Study 4 summary: the tuned, converged comparison: every arm's PP and PS heads at its own validation-selected learning rate, cap 200 epochs, five seeds. Intervals from the hierarchical bootstrap over 8 matches and 5 seeds ($B=10{,}000$); family of 50 hypotheses. Reading conventions: Sec.~\ref{par:reading}. Full family: Supplement Table~\suppref{tab:tuned}}
\label{tab:summary-tuned}
\centering
\footnotesize
\setlength{\tabcolsep}{2pt}
\renewcommand{\arraystretch}{1.0}
\begin{tabular}{lrrrr}
\toprule
Pair & Selection NLL\,$\downarrow$ & Completion log-loss\,$\downarrow$ & Mass $<$5\,m\,$\uparrow$ & Argmax dist.\ (m)\,$\downarrow$ \\
\midrule
\Gtwo{}$-$C & $-0.292$ $[-0.338,-0.245]$ & $-0.013$ $[-0.022,-0.007]$ & $+0.038$ $[+0.031,+0.045]$ & $-0.87$ $[-1.17,-0.60]$ \\
{\scriptsize\itshape matches, adj.\ $p$} & {\scriptsize sd 0.024, 8/8, $p<0.01$ $\ast$$\circ$} & {\scriptsize sd 0.009, 8/8, $p<0.01$ $\ast$$\circ$} & {\scriptsize sd 0.005, 8/8, $p<0.01$ $\ast$$\circ$} & {\scriptsize sd 0.29, 8/8, $p<0.01$ $\ast$$\circ$} \\[1pt]
\Gtwo{}$-$original & $-0.319$ $[-0.364,-0.275]$ & $-0.035$ $[-0.042,-0.029]$ & $+0.033$ $[+0.026,+0.042]$ & $-1.36$ $[-1.64,-1.09]$ \\
{\scriptsize\itshape matches, adj.\ $p$} & {\scriptsize sd 0.015, 8/8, $p<0.01$ $\ast$$\circ$} & {\scriptsize sd 0.004, 8/8, $p<0.01$ $\ast$$\circ$} & {\scriptsize sd 0.008, 8/8, $p<0.01$ $\ast$$\circ$} & {\scriptsize sd 0.28, 8/8, $p<0.01$ $\ast$$\circ$} \\[1pt]
\Gtwo{}$-$wide & $-0.328$ $[-0.368,-0.291]$ & $-0.031$ $[-0.039,-0.023]$ & $+0.035$ $[+0.027,+0.046]$ & $-1.39$ $[-1.72,-1.06]$ \\
{\scriptsize\itshape matches, adj.\ $p$} & {\scriptsize sd 0.026, 8/8, $p<0.01$ $\ast$$\circ$} & {\scriptsize sd 0.008, 8/8, $p<0.01$ $\ast$$\circ$} & {\scriptsize sd 0.010, 8/8, $p<0.01$ $\ast$$\circ$} & {\scriptsize sd 0.36, 8/8, $p<0.01$ $\ast$$\circ$} \\[1pt]
C$-$original & $-0.027$ $[-0.063,+0.006]$ & $-0.022$ $[-0.031,-0.012]$ & $-0.005$ $[-0.010,+0.000]$ & $-0.49$ $[-0.74,-0.18]$ \\
{\scriptsize\itshape matches, adj.\ $p$} & {\scriptsize sd 0.035, 7/8, $p=0.297$} & {\scriptsize sd 0.007, 8/8, $p<0.01$ $\ast$$\circ$} & {\scriptsize sd 0.005, 2/8, $p=0.320$} & {\scriptsize sd 0.29, 8/8, $p<0.01$ $\ast$$\circ$} \\[1pt]
C$-$wide & $-0.036$ $[-0.069,+0.007]$ & $-0.018$ $[-0.027,-0.009]$ & $-0.002$ $[-0.011,+0.005]$ & $-0.52$ $[-0.72,-0.28]$ \\
{\scriptsize\itshape matches, adj.\ $p$} & {\scriptsize sd 0.043, 8/8, $p=0.094$} & {\scriptsize sd 0.009, 8/8, $p<0.01$ $\ast$$\circ$} & {\scriptsize sd 0.009, 3/8, $p=0.656$} & {\scriptsize sd 0.20, 8/8, $p<0.01$ $\ast$$\circ$} \\
\bottomrule
\end{tabular}
\end{table}

\begin{table}[t]
\caption{Study 12 (B) summary: the raw-feature arm D$_{\text{raw}}$ under the Study 4 protocol, its own validation-selected learning rate, cap 200 epochs, five seeds, against the locked Study 4 runs of \Gtwo{} and C. Intervals from the hierarchical bootstrap over 8 matches and 5 seeds ($B=10{,}000$); family of 20 hypotheses. Reading conventions: Sec.~\ref{par:reading}. Full family: Supplement Table~\suppref{tab:tuned-raw}}
\label{tab:summary-tuned-raw}
\centering
\footnotesize
\setlength{\tabcolsep}{2pt}
\renewcommand{\arraystretch}{1.0}
\begin{tabular}{lrrrr}
\toprule
Pair & Selection NLL\,$\downarrow$ & Completion log-loss\,$\downarrow$ & Mass $<$5\,m\,$\uparrow$ & Argmax dist.\ (m)\,$\downarrow$ \\
\midrule
D$_{\text{raw}}$ $-$ C & $-0.225$ $[-0.256,-0.195]$ & $-0.009$ $[-0.016,-0.002]$ & $+0.027$ $[+0.021,+0.033]$ & $-0.64$ $[-0.82,-0.50]$ \\
{\scriptsize\itshape matches, adj.\ $p$} & {\scriptsize sd 0.019, 8/8, $p<0.01$ $\ast$$\circ$} & {\scriptsize sd 0.005, 7/8, $p=0.125$} & {\scriptsize sd 0.006, 8/8, $p<0.01$ $\ast$$\circ$} & {\scriptsize sd 0.14, 8/8, $p<0.01$ $\ast$$\circ$} \\[1pt]
\Gtwo{} $-$ D$_{\text{raw}}$ & $-0.067$ $[-0.088,-0.045]$ & $-0.005$ $[-0.010,+0.001]$ & $+0.011$ $[+0.008,+0.015]$ & $-0.23$ $[-0.50,+0.00]$ \\
{\scriptsize\itshape matches, adj.\ $p$} & {\scriptsize sd 0.010, 8/8, $p<0.01$ $\ast$$\circ$} & {\scriptsize sd 0.005, 6/8, $p=0.352$} & {\scriptsize sd 0.004, 8/8, $p<0.01$ $\ast$$\circ$} & {\scriptsize sd 0.27, 7/8, $p=0.047$ $\circ$} \\
\bottomrule
\end{tabular}
\end{table}

\subsection{What the painted front end supplies}
\label{sec:frontend}

The three studies above measure what the painted front end is worth; the two
here, both on the reused eight-match holdout, take it apart. The attribution
ladder prices the front end as a package and separates the painted channels
from message passing; the ablations then vary one element at a time.

\subsubsection{Attribution under the pre-specified replication}
\label{sec:attribution}

\begin{figure}[tbp]
\centering
\includegraphics[width=\textwidth]{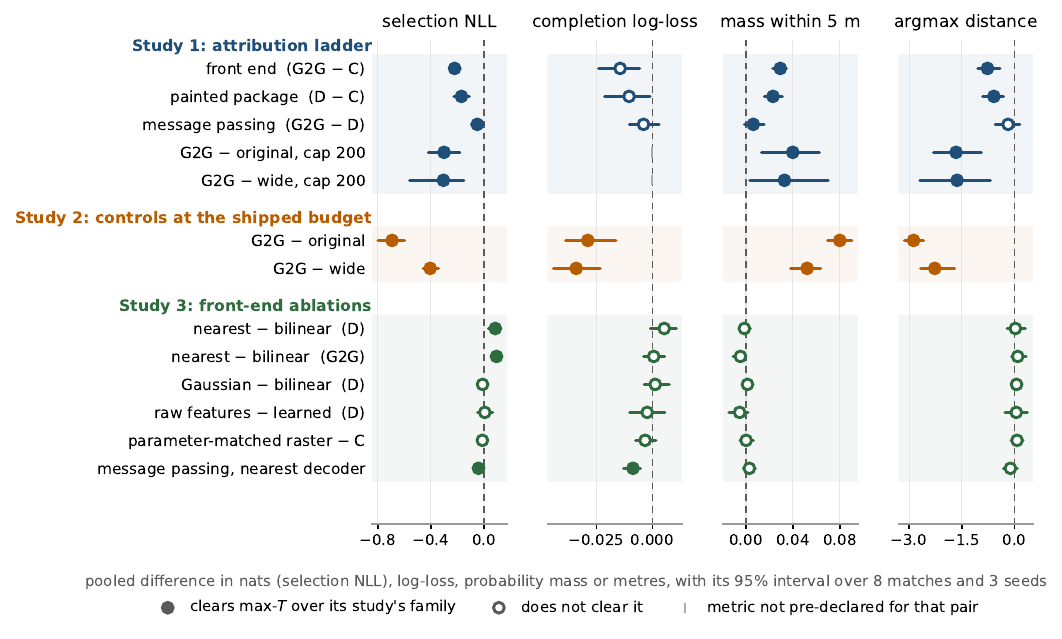}
\caption{Multi-seed effects with 95\% hierarchical-bootstrap intervals over 8
matches and 3 seeds, for the four headline metrics, with one colour and one
band per study: the Study 1 attribution ladder (front end, painted package,
message passing) with its cap-200 convergence pairs, the Study 2 multi-seed
controls at the shipped budget, and the Study 3 front-end ablations
(Sec.~\ref{sec:ablation}). A filled marker means the contrast clears the
exact sign-flip max-$T$ test over its study's pre-declared family at
$\alpha=0.05$, the test this paper prioritises; a hollow marker means it does
not. A vertical tick at zero marks a metric that was not pre-declared for
that pair, the cap-200 reruns having trained the selection head only}
\label{fig:forest}
\end{figure}

The painted front end earns the improvement. Adding relational reasoning on
top of it moves the selection score by $-0.0515$ nats, an estimate whose
nominal interval excludes zero but whose per-seed spread is as large as the
effect. Fig.~\ref{fig:forest} shows every headline effect of the three multi-seed
studies it covers, with its interval.

The replication is Study 1 of Sec.~\ref{sec:prespec}: we trained three fresh
seeds each of \Gtwo{}, arm~C and arm~D, plus cap-200 selection-head reruns of
both controls, on the same eight matches and split signature. Arm~C and arm~D
differ only in the painted front end; arm~D and \Gtwo{} differ only in the
per-node encoder, which \Gtwo{} replaces with the relational one.
Table~\ref{tab:confirm} prices the three design steps on the four headline
metrics; Supplement Tables~\suppref{tab:hier-attrib}
and~\suppref{tab:hier-conv} print all 54 hypotheses under both procedures.
The exact sign-flip test applies here in a restricted sense: the
pre-specified recipe averages each within-match delta over the seeds
\emph{before} the sign flip, so the test says nothing about seed
variability, and the hierarchical bootstrap, which resamples seeds as well,
is the stricter procedure on that axis (Sec.~\ref{sec:prespec}).

Adding the painted front end to \vtwo{} (arm~C $\to$~D), the \emph{painted
package} of $29{,}580$ parameters, is worth $-0.1705$ nats of selection NLL;
the whole front end including message passing (arm~C $\to$ \Gtwo{}) is worth
$-0.2220$ (Table~\ref{tab:confirm}), so the point estimates put the painted
package at roughly three-quarters of the front-end selection improvement.
Because the painted front end bundles the per-node network, the entity
features, the bilinear projection, the 36 extra channels and the wider input
stem, this step is evidence for the composite package, not for splatting or
sub-cell geometry in isolation; Secs.~\ref{sec:ablation}
and~\ref{sec:raster} separate them. On selection NLL, local mass and argmax
distance both the whole front end and the painted package clear the max-$T$
test, ahead in all eight matches; on completion log-loss neither does,
although both intervals exclude zero. Message passing on top of the package
clears max-$T$ on selection NLL and local mass and on nothing else, so on this holdout the relational increment is small and seed-sensitive rather than absent: across the three seeds (Supplement Table~\suppref{tab:perseed})
the \Gtwo{}$-$D selection-NLL difference ranges from nearly a tenth of a nat
to nearly nothing, a spread as large as the effect. The combined learned
increment, encoder and message passing together, is resolved over 64
matches and under five-seed tuning (Secs.~\ref{sec:cv}, \ref{sec:tuned}).

\begin{table}[t]
\caption{Study 1 summary: which design step earns the improvement: \emph{front end} is the whole graph front end added to \vtwo{} (\Gtwo{}$-$C); \emph{painted package} is that front end without relational message passing, i.e.\ the per-node MLP, splat operator, 36 painted channels and wider input stem together (D$-$C); \emph{message passing} is the increment message passing adds on top of the package (\Gtwo{}$-$D). Three fresh seeds per arm on the eight-match holdout. Intervals from the hierarchical bootstrap over 8 matches and 3 seeds ($B=10{,}000$); family of 54 hypotheses. Reading conventions: Sec.~\ref{par:reading}. Full family: Supplement Tables~\suppref{tab:hier-attrib} and~\suppref{tab:hier-conv}}
\label{tab:confirm}
\centering
\footnotesize
\setlength{\tabcolsep}{2pt}
\renewcommand{\arraystretch}{1.0}
\begin{tabular}{lrrrr}
\toprule
Pair & Selection NLL\,$\downarrow$ & Completion log-loss\,$\downarrow$ & Mass $<$5\,m\,$\uparrow$ & Argmax dist.\ (m)\,$\downarrow$ \\
\midrule
\Gtwo{}$-$C & $-0.222$ $[-0.259,-0.181]$ & $-0.014$ $[-0.024,-0.006]$ & $+0.029$ $[+0.023,+0.034]$ & $-0.77$ $[-1.04,-0.41]$ \\
{\scriptsize\itshape matches, adj.\ $p$} & {\scriptsize sd 0.020, 8/8, $p<0.01$ $\ast$$\circ$} & {\scriptsize sd 0.007, 8/8, $p=0.070$ $\ast$} & {\scriptsize sd 0.005, 8/8, $p<0.01$ $\ast$$\circ$} & {\scriptsize sd 0.32, 8/8, $p<0.01$ $\ast$$\circ$} \\[1pt]
D$-$C & $-0.171$ $[-0.227,-0.112]$ & $-0.010$ $[-0.021,-0.001]$ & $+0.023$ $[+0.016,+0.031]$ & $-0.59$ $[-0.90,-0.31]$ \\
{\scriptsize\itshape matches, adj.\ $p$} & {\scriptsize sd 0.046, 8/8, $p<0.01$ $\ast$$\circ$} & {\scriptsize sd 0.010, 8/8, $p=0.055$} & {\scriptsize sd 0.007, 8/8, $p<0.01$ $\ast$$\circ$} & {\scriptsize sd 0.29, 8/8, $p<0.01$ $\ast$$\circ$} \\[1pt]
\Gtwo{}$-$D & $-0.051$ $[-0.095,-0.005]$ & $-0.004$ $[-0.010,+0.003]$ & $+0.006$ $[-0.001,+0.015]$ & $-0.18$ $[-0.55,+0.15]$ \\
{\scriptsize\itshape matches, adj.\ $p$} & {\scriptsize sd 0.046, 8/8, $p=0.016$ $\circ$} & {\scriptsize sd 0.005, 6/8, $p=0.516$} & {\scriptsize sd 0.009, 8/8, $p<0.01$ $\circ$} & {\scriptsize sd 0.35, 8/8, $p=0.086$} \\
\bottomrule
\end{tabular}
\end{table}

\subsubsection{Front-end ablations}
\label{sec:ablation}

How the write is placed carries about half the painted package's selection
gain; for what is written, how it is anti-aliased and how many parameters the
core has, no difference is detected under this protocol. Study 3
(Sec.~\ref{sec:prespec}) takes apart the package that the attribution ladder
priced, with trained ablations, three seeds each, against the locked Study 1
runs of D, \Gtwo{} and C, each ablation changing one element of the design
and holding the rest fixed. Table~\ref{tab:summary-ablation} summarises the
family and Supplement Table~\suppref{tab:ablation} gives all 60 hypotheses.

\paragraph{Bilinear placement carries about half the package.}
Replacing bilinear painting by nearest-cell painting discards the fractional
position at which each write is placed, and changes the interpolation with
it; the node features themselves are unchanged. It costs the per-node arm and
\Gtwo{} each close to a tenth of a nat of selection NLL, both contrasts
clearing the max-$T$ test, and costs \Gtwo{} a little local mass. Against the
painted package's $-0.1705$ nats (Table~\ref{tab:confirm}), roughly half the
selection gain is therefore attributable to bilinear placement at continuous
coordinates, a point estimate for a change that alters placement and
interpolation together. Completion is untouched, which matches the
inference-time quantisation probe of Sec.~\ref{sec:probes}: sub-cell
placement acts on selection.

\paragraph{No effect of anti-aliasing is detected.}
A normalised $3\times3$ Gaussian painting of half-cell width does not
separate from bilinear painting on any headline metric under this protocol.
Its selection interval is narrow, so any difference between the two kernels
is at most a few hundredths of a nat, well below the effect of bilinear
placement itself.

\paragraph{Raw features and the learned embedding are not separated on the holdout.}
Painting the nine raw node features bilinearly with no learned encoder
(D$_{\text{raw}}$, 27 channels instead of 36) does not separate from the
per-node MLP arm on any headline metric, and the two arms' seed-averaged
selection NLLs lie within a hundredth of a nat of each other. This is not
evidence of equivalence: the selection interval reaches five to six hundredths
of a nat in either direction, as wide as the learned increment that the
cross-validation resolves (Sec.~\ref{sec:cv}). On this holdout, most of what
the painted package supplies is available from painting the players' own
features at continuous positions, and the learned embedding adds no
detectable further gain.

\paragraph{Parameters do not explain it.}
Widening the raster core to within $0.6\%$ of \Gtwo{}'s parameter count
moves arm~C by about a hundredth of a nat on selection NLL, a null contrast,
against the fifth of a nat the whole front end adds to the same core. The
front-end gain is not a parameter-count effect.

\paragraph{Message passing under the nearest decoder.}
With sub-cell placement removed from both arms, adding the two
message-passing blocks is worth a few hundredths of a nat of selection NLL
and under a hundredth of completion log-loss, both clearing the max-$T$ test
and both of the order of what message passing adds on top of bilinear
painting (Table~\ref{tab:confirm}). The relational encoder recovers a little
of what nearest-cell painting loses, but only a little:
\Gtwo{}$_{\text{nearest}}$ remains behind the non-relational bilinear arm~D.
Taken together with Sec.~\ref{sec:attribution}, these holdout studies locate
most of the gain the front end adds to \vtwo{} in writing per-player
information onto the grid at continuous coordinates. No separate
contribution of the learned embedding, the anti-aliasing kernel or the
parameter count is detected here, and relational message passing adds a
small increment of the order of its own seed variation under either decoder.
The larger cross-validation and tuned comparisons resolve the combined
contribution of the learned encoder and message passing, but not either one
alone.

\begin{table}[t]
\caption{Study 3 summary: front-end ablations, three seeds each, against the locked Study 1 runs of D, \Gtwo{} and C: nearest-cell and Gaussian painting against bilinear, the nine raw node features against the learned embedding, a parameter-matched raster core, and message passing under the nearest-cell decoder. Intervals from the hierarchical bootstrap over 8 matches and 3 seeds ($B=10{,}000$); family of 60 hypotheses. Reading conventions: Sec.~\ref{par:reading}. Full family: Supplement Table~\suppref{tab:ablation}}
\label{tab:summary-ablation}
\centering
\scriptsize
\setlength{\tabcolsep}{2pt}
\renewcommand{\arraystretch}{1.0}
\begin{tabular}{lrrrr}
\toprule
Pair & Selection NLL\,$\downarrow$ & Completion log-loss\,$\downarrow$ & Mass $<$5\,m\,$\uparrow$ & Argmax dist.\ (m)\,$\downarrow$ \\
\midrule
nearest $-$ bilinear (D) & $+0.084$ $[+0.033,+0.125]$ & $+0.005$ $[-0.001,+0.011]$ & $-0.001$ $[-0.006,+0.004]$ & $+0.03$ $[-0.20,+0.31]$ \\
{\scriptsize\itshape matches, adj.\ $p$} & {\scriptsize sd 0.046, 0/8, $p<0.01$ $\circ$} & {\scriptsize sd 0.005, 1/8, $p=0.211$} & {\scriptsize sd 0.004, 2/8, $p=0.984$} & {\scriptsize sd 0.22, 4/8, $p=1$} \\[1pt]
Gaussian $-$ bilinear (D) & $-0.012$ $[-0.024,+0.001]$ & $+0.001$ $[-0.004,+0.007]$ & $+0.001$ $[-0.001,+0.004]$ & $+0.06$ $[-0.05,+0.19]$ \\
{\scriptsize\itshape matches, adj.\ $p$} & {\scriptsize sd 0.012, 7/8, $p=0.078$} & {\scriptsize sd 0.006, 3/8, $p=1$} & {\scriptsize sd 0.002, 6/8, $p=0.852$} & {\scriptsize sd 0.06, 2/8, $p=1$} \\[1pt]
raw feat.\ $-$ learned (D) & $+0.005$ $[-0.049,+0.064]$ & $-0.002$ $[-0.010,+0.006]$ & $-0.005$ $[-0.014,+0.002]$ & $+0.05$ $[-0.27,+0.36]$ \\
{\scriptsize\itshape matches, adj.\ $p$} & {\scriptsize sd 0.058, 3/8, $p=1$} & {\scriptsize sd 0.007, 5/8, $p=1$} & {\scriptsize sd 0.008, 2/8, $p=0.438$} & {\scriptsize sd 0.28, 4/8, $p=1$} \\[1pt]
C$_{\text{matched}}$ $-$ C & $-0.013$ $[-0.042,+0.012]$ & $-0.003$ $[-0.008,+0.002]$ & $+0.000$ $[-0.005,+0.006]$ & $+0.07$ $[-0.09,+0.24]$ \\
{\scriptsize\itshape matches, adj.\ $p$} & {\scriptsize sd 0.016, 4/8, $p=1$} & {\scriptsize sd 0.001, 6/8, $p=0.945$} & {\scriptsize sd 0.005, 4/8, $p=1$} & {\scriptsize sd 0.06, 3/8, $p=1$} \\[1pt]
nearest $-$ bilinear (\Gtwo{}) & $+0.093$ $[+0.072,+0.117]$ & $+0.001$ $[-0.004,+0.005]$ & $-0.005$ $[-0.011,-0.000]$ & $+0.09$ $[-0.08,+0.33]$ \\
{\scriptsize\itshape matches, adj.\ $p$} & {\scriptsize sd 0.015, 0/8, $p<0.01$ $\ast$$\circ$} & {\scriptsize sd 0.003, 3/8, $p=1$} & {\scriptsize sd 0.005, 0/8, $p=0.094$} & {\scriptsize sd 0.19, 3/8, $p=0.969$} \\[1pt]
msg.\ pass.\ (nearest) & $-0.042$ $[-0.080,-0.011]$ & $-0.009$ $[-0.013,-0.005]$ & $+0.003$ $[-0.002,+0.008]$ & $-0.12$ $[-0.31,+0.08]$ \\
{\scriptsize\itshape matches, adj.\ $p$} & {\scriptsize sd 0.035, 8/8, $p<0.01$ $\circ$} & {\scriptsize sd 0.001, 8/8, $p<0.01$ $\ast$$\circ$} & {\scriptsize sd 0.004, 7/8, $p=0.734$} & {\scriptsize sd 0.17, 6/8, $p=0.852$} \\
\bottomrule
\end{tabular}
\end{table}

\subsection{Raster controls with sub-cell information}
\label{sec:raster}

Neither sub-cell offset channels nor a raster four times finer closes the gap
to the painted front end, and the same splat on the original SoccerMap core is
worth $-0.295$ nats. The ablations of Sec.~\ref{sec:ablation} and the
interface alternatives of Sec.~\ref{sec:setattn} vary the painted front end
and hold the raster fixed; two objections run the other way. The first is
that a raster core loses to a painted front end only because hard binning
discards where inside a cell a player stands, so that a raster carrying that
information would close the gap without any graph; the second is that the
original SoccerMap core was never given the painted channels, so that the
margin over it conflates core and front end. Studies 8 and 9
(Sec.~\ref{sec:prespec}) test specific versions of these alternatives.

\paragraph{Design: sub-cell information on the raster.}
Study 8 registers two arms on the untouched engine. C$_{\text{offsets}}$ is
arm~C's \vtwo{} core on the 13 raster channels plus six deterministic
sub-cell mean-offset channels, two per node type, each cell carrying the
mean offset from its centre of the nodes of that type inside it; the
channels hold no learned parameter and are the arm's only difference from
arm~C. Original$_{\text{splat}}$ is the original three-scale SoccerMap core
on the 13 raster channels plus the 27 bilinearly painted raw node features
of D$_{\text{raw}}$ (Sec.~\ref{sec:ablation}); the painted channels are its
only difference from the original control, and it has fewer parameters than
the wide control. Each arm trains PP and PS heads at the shipped recipe
under the three Study 1 seeds and is compared with the locked runs of C, D,
original, wide and D$_{\text{raw}}$, none retrained: five pre-declared pairs, 50 hypotheses, with the hierarchical bootstrap
over 8 matches and 3 seeds and the exact 256-pattern max-$T$ test;
Table~\ref{tab:summary-raster} summarises the family and Supplement
Table~\suppref{tab:raster} gives every hypothesis.

\paragraph{Sub-cell offsets on the raster core.}
Giving the raster core the sub-cell position of every player and the ball
helps it, but not much (Table~\ref{tab:summary-raster}). C$_{\text{offsets}}$
improves on arm~C by a few hundredths of a nat of selection NLL and by less
than a hundredth of completion log-loss; the completion contrast clears the
max-$T$ test, the selection contrast does not, and local mass and
argmax distance are null. With seed-averaged selection NLLs of $5.446$ for
C$_{\text{offsets}}$ against $5.490$ for C and $5.319$ for D, the offset
channels close about a quarter of the gap between the raster core and the
painted front end, and none of the gap in local mass: sub-cell position
written as an explicit raster feature is worth a fraction of what the same
position is worth as the place the embedding is written.

\paragraph{Learned and raw painting against the offset raster.}
Against the raster that carries the same sub-cell position, both painted
arms stay ahead by the same margin: D and D$_{\text{raw}}$ each lead
C$_{\text{offsets}}$ by about an eighth of a nat of selection NLL, with more
local mass and a closer argmax, all under both procedures, while completion
is level; every C$_{\text{offsets}}$ head early-stopped, so the budget was
not binding. What the splat adds over an offset raster is therefore not the
position itself, which both arms have, but writing the player's own features
at that position, and the raw-feature painter with no learned parameters
recovers essentially all of it, as it did against arm~C in
Sec.~\ref{sec:ablation}.

\paragraph{The painted front end on the original core.}
The painted front end is worth more on the original core than on ours.
Original$_{\text{splat}}$ improves on the original control by $-0.295$ nats
of selection NLL, with completion, local mass and argmax distance improving
as well, every contrast under both procedures, where the same raw-feature write adds $-0.165$ nats to the \vtwo{} core
(D$_{\text{raw}}$ against C on the locked runs, as below); against the
wide control, which has more than twice its parameters, it is level on
selection and local mass and ahead on completion under both procedures, at
equal budget. The margin the paper reports over the original SoccerMap
therefore does not conflate core and front end in the direction a reader
might fear, the front end alone on the original core being worth close to
half of the matched-budget gap, and the entity-to-grid interface is not
tied to the \vtwo{} core.

\begin{table}[t]
\caption{Study 8 summary: the raster controls with sub-cell information: the raster core of arm~C plus six deterministic sub-cell mean-offset channels (C$_{\text{offsets}}$) and the original SoccerMap core plus the 27 painted raw node features (original$_{\text{splat}}$), three seeds each, against the locked runs of C, D, original, wide and D$_{\text{raw}}$. Intervals from the hierarchical bootstrap over 8 matches and 3 seeds ($B=10{,}000$); family of 50 hypotheses. Reading conventions: Sec.~\ref{par:reading}. Full family: Supplement Table~\suppref{tab:raster}}
\label{tab:summary-raster}
\centering
\scriptsize
\setlength{\tabcolsep}{2pt}
\renewcommand{\arraystretch}{1.0}
\begin{tabular}{lrrrr}
\toprule
Pair & Selection NLL\,$\downarrow$ & Completion log-loss\,$\downarrow$ & Mass $<$5\,m\,$\uparrow$ & Argmax dist.\ (m)\,$\downarrow$ \\
\midrule
C$_{\text{offsets}}$ $-$ C & $-0.044$ $[-0.075,-0.011]$ & $-0.006$ $[-0.010,-0.003]$ & $-0.002$ $[-0.005,+0.001]$ & $+0.28$ $[-0.06,+0.59]$ \\
{\scriptsize\itshape matches, adj.\ $p$} & {\scriptsize sd 0.001, 6/8, $p=0.406$} & {\scriptsize sd 0.000, 8/8, $p=0.016$ $\circ$} & {\scriptsize sd 0.002, 2/8, $p=0.844$} & {\scriptsize sd 0.30, 1/8, $p=0.125$} \\[1pt]
D$_{\text{raw}}$ $-$ C$_{\text{offsets}}$ & $-0.122$ $[-0.152,-0.090]$ & $-0.006$ $[-0.011,-0.002]$ & $+0.019$ $[+0.014,+0.026]$ & $-0.81$ $[-1.09,-0.55]$ \\
{\scriptsize\itshape matches, adj.\ $p$} & {\scriptsize sd 0.027, 8/8, $p<0.01$ $\ast$$\circ$} & {\scriptsize sd 0.003, 7/8, $p=0.102$} & {\scriptsize sd 0.005, 8/8, $p<0.01$ $\ast$$\circ$} & {\scriptsize sd 0.22, 8/8, $p<0.01$ $\ast$$\circ$} \\[1pt]
D $-$ C$_{\text{offsets}}$ & $-0.127$ $[-0.172,-0.078]$ & $-0.004$ $[-0.015,+0.006]$ & $+0.025$ $[+0.016,+0.033]$ & $-0.86$ $[-1.32,-0.36]$ \\
{\scriptsize\itshape matches, adj.\ $p$} & {\scriptsize sd 0.046, 8/8, $p<0.01$ $\ast$$\circ$} & {\scriptsize sd 0.010, 7/8, $p=0.430$} & {\scriptsize sd 0.008, 8/8, $p<0.01$ $\ast$$\circ$} & {\scriptsize sd 0.49, 8/8, $p<0.01$ $\ast$$\circ$} \\[1pt]
original$_{\text{splat}}$ $-$ original & $-0.295$ $[-0.416,-0.214]$ & $-0.017$ $[-0.024,-0.008]$ & $+0.026$ $[+0.020,+0.032]$ & $-0.64$ $[-0.81,-0.43]$ \\
{\scriptsize\itshape matches, adj.\ $p$} & {\scriptsize sd 0.108, 8/8, $p<0.01$ $\ast$$\circ$} & {\scriptsize sd 0.008, 8/8, $p<0.01$ $\ast$$\circ$} & {\scriptsize sd 0.005, 8/8, $p<0.01$ $\ast$$\circ$} & {\scriptsize sd 0.15, 8/8, $p<0.01$ $\ast$$\circ$} \\[1pt]
original$_{\text{splat}}$ $-$ wide & $-0.009$ $[-0.071,+0.044]$ & $-0.022$ $[-0.030,-0.015]$ & $-0.002$ $[-0.010,+0.008]$ & $-0.03$ $[-0.35,+0.31]$ \\
{\scriptsize\itshape matches, adj.\ $p$} & {\scriptsize sd 0.057, 5/8, $p=1$} & {\scriptsize sd 0.006, 8/8, $p<0.01$ $\ast$$\circ$} & {\scriptsize sd 0.009, 2/8, $p=0.805$} & {\scriptsize sd 0.31, 4/8, $p=1$} \\
\bottomrule
\end{tabular}
\end{table}

\paragraph{Design: the half-metre raster.}
Study 9 trains the same two cores on a raster four times finer:
C$_{\text{half}}$ and original$_{\text{half}}$ apply the \vtwo{} core of
arm~C and the original SoccerMap core to the 13 deterministic channels
rasterised at $0.5$\,m, $136\times208=28{,}288$ cells instead of
$68\times104=7{,}072$; the convolutions are resolution-free, so the grid is
the only difference from those arms. We built the $0.5$\,m cache from the
same locked record cache by the unchanged extractor, pinned its digest in
the commitment and self-checked it against the locked $1$\,m cache. Every
$0.5$\,m arm is scored on the $1$\,m grid by a rule fixed before training
and quoted verbatim in Supplement Sec.~\suppref{supp:halfrule}: tempered
$0.5$\,m selection probabilities are sum-pooled $2\times2$ to the $1$\,m
cells, and the completion probability of a pass is the tempered sigmoid at
the $0.5$\,m cell containing the observed endpoint. Each arm trains under
the three Study 1 seeds and is compared with the locked runs of C, D,
\Gtwo{} and original: four pre-declared pairs, 40 hypotheses, with the same bootstrap and
test; Table~\ref{tab:summary-halfgrid} summarises the family and Supplement
Table~\suppref{tab:halfgrid} gives every hypothesis.

\paragraph{A finer raster.}
A raster four times finer does not close the gap; on two of three selection
metrics it widens it (Table~\ref{tab:summary-halfgrid}). C$_{\text{half}}$
moves the \vtwo{} core's selection NLL by a few hundredths of a nat, an
interval that touches zero and a contrast far from the adjusted max-$T$ threshold,
while its argmax lands nearly a metre farther from the observed endpoint and
its mass within 5\,m falls, both under both procedures; completion is
unchanged. The original core is simply worse at half a metre, a fifth of a
nat behind on selection NLL and behind on local mass. Given four times as
many cells to put its mass in, a raster core spreads it: the likelihood at
the observed cell changes little, and the surface's peak and its local
concentration get worse.

\paragraph{The painted front end against the finer raster.}
Against the half-metre raster the painted arms keep the margin they hold
against the metre raster: D and \Gtwo{} lead C$_{\text{half}}$ on every
headline metric in all eight matches under both procedures. Read together
with Study 8, the two objections are answered in the same way, within the
reach of the controls run: give the raster the sub-cell position as a
feature and it recovers a quarter of the painted front end's gain; give it
cells four times smaller and its selection likelihood shows no statistically
resolved improvement while its localisation worsens. The finer raster keeps
the convolutional architecture, so its physical receptive field halves with
the cell; the study rules out resolution as a substitute for the write under
that architecture, not resolution in general. What the splat adds, on this
evidence, is neither resolution nor position alone: it is each player's own
features written at that player's position, in a form the convolution reads
directly.

\begin{table}[t]
\caption{Study 9 summary: the half-metre raster: the raster core of arm~C (C$_{\text{half}}$) and the original SoccerMap core (original$_{\text{half}}$) on the 13 channels rasterised at 0.5\,m, scored for comparability with the 1\,m arms, three seeds each, against the locked runs of C, D, \Gtwo{} and original. Intervals from the hierarchical bootstrap over 8 matches and 3 seeds ($B=10{,}000$); family of 40 hypotheses. Reading conventions: Sec.~\ref{par:reading}. Full family: Supplement Table~\suppref{tab:halfgrid}}
\label{tab:summary-halfgrid}
\centering
\scriptsize
\setlength{\tabcolsep}{2pt}
\renewcommand{\arraystretch}{1.0}
\begin{tabular}{lrrrr}
\toprule
Pair & Selection NLL\,$\downarrow$ & Completion log-loss\,$\downarrow$ & Mass $<$5\,m\,$\uparrow$ & Argmax dist.\ (m)\,$\downarrow$ \\
\midrule
C$_{\text{half}}$ $-$ C & $-0.047$ $[-0.106,+0.001]$ & $+0.007$ $[-0.003,+0.015]$ & $-0.011$ $[-0.016,-0.005]$ & $+0.93$ $[+0.70,+1.26]$ \\
{\scriptsize\itshape matches, adj.\ $p$} & {\scriptsize sd 0.034, 5/8, $p=0.609$} & {\scriptsize sd 0.007, 1/8, $p=0.461$} & {\scriptsize sd 0.005, 0/8, $p=0.016$ $\ast$$\circ$} & {\scriptsize sd 0.18, 0/8, $p<0.01$ $\ast$$\circ$} \\[1pt]
original$_{\text{half}}$ $-$ original & $+0.199$ $[+0.082,+0.274]$ & $-0.000$ $[-0.009,+0.010]$ & $-0.019$ $[-0.022,-0.015]$ & $+0.28$ $[+0.09,+0.47]$ \\
{\scriptsize\itshape matches, adj.\ $p$} & {\scriptsize sd 0.104, 0/8, $p<0.01$ $\ast$$\circ$} & {\scriptsize sd 0.004, 3/8, $p=1$} & {\scriptsize sd 0.003, 0/8, $p<0.01$ $\ast$$\circ$} & {\scriptsize sd 0.15, 0/8, $p=0.023$ $\circ$} \\[1pt]
D $-$ C$_{\text{half}}$ & $-0.123$ $[-0.167,-0.087]$ & $-0.018$ $[-0.026,-0.009]$ & $+0.034$ $[+0.028,+0.040]$ & $-1.52$ $[-1.91,-1.20]$ \\
{\scriptsize\itshape matches, adj.\ $p$} & {\scriptsize sd 0.033, 8/8, $p<0.01$ $\ast$$\circ$} & {\scriptsize sd 0.004, 8/8, $p<0.01$ $\ast$$\circ$} & {\scriptsize sd 0.005, 8/8, $p<0.01$ $\ast$$\circ$} & {\scriptsize sd 0.30, 8/8, $p<0.01$ $\ast$$\circ$} \\[1pt]
\Gtwo{} $-$ C$_{\text{half}}$ & $-0.175$ $[-0.224,-0.121]$ & $-0.022$ $[-0.031,-0.015]$ & $+0.041$ $[+0.035,+0.046]$ & $-1.70$ $[-1.96,-1.43]$ \\
{\scriptsize\itshape matches, adj.\ $p$} & {\scriptsize sd 0.048, 8/8, $p<0.01$ $\ast$$\circ$} & {\scriptsize sd 0.006, 8/8, $p<0.01$ $\ast$$\circ$} & {\scriptsize sd 0.004, 8/8, $p<0.01$ $\ast$$\circ$} & {\scriptsize sd 0.14, 8/8, $p<0.01$ $\ast$$\circ$} \\
\bottomrule
\end{tabular}
\end{table}

\paragraph{The interface on a canonical U-Net core.}
The gain is not a property of \vtwo{} either. Study 13
(Sec.~\ref{sec:prespec}) trains a canonical three-level 2-D
U-Net~\citep{ronneberger2015unet}, parameter-matched to arm~C within one
percent, on the raster channels alone (U$_{\text{raster}}$) and with the 27
painted raw node features of D$_{\text{raw}}$ added (U$_{\text{raw}}$),
three seeds each at the shipped recipe, against the locked runs of C,
\Gtwo{} and D$_{\text{raw}}$; Table~\ref{tab:summary-unet} summarises the
family and Supplement Table~\suppref{tab:unet} gives all 40 hypotheses.
Adding the write to the U-Net core is worth $-0.242$ nats of selection NLL
$[-0.264,-0.213]$, ahead in all eight matches under both procedures, and
completion, local mass and the argmax move with it; on \vtwo{} the same
write, D$_{\text{raw}}$ against C from the locked runs, is worth $-0.165$.
The write therefore pays on three cores: the published SoccerMap, \vtwo{}
and a canonical U-Net. The U-Net is the weaker core: raster-only it trails
arm~C by $+0.366$ nats of selection NLL and painted it trails
D$_{\text{raw}}$ by $+0.290$, behind in every match under both procedures,
so the modernised core is the stronger reader at this shared recipe.
This core comparison was not repeated under per-arm learning-rate tuning.

\begin{table}[t]
\caption{Study 13 summary: the interface on a canonical U-Net core: U$_{\text{raster}}$ (a three-level 2-D U-Net on the 13 raster channels, parameter-matched to arm~C within 1\%) and U$_{\text{raw}}$ (the same U-Net plus the 27 painted raw node features of D$_{\text{raw}}$), three seeds each, against the locked runs of C and \Gtwo{} (Study 1) and D$_{\text{raw}}$ (Study 3). Intervals from the hierarchical bootstrap over 8 matches and 3 seeds ($B=10{,}000$); family of 40 hypotheses. Reading conventions: Sec.~\ref{par:reading}. Full family: Supplement Table~\suppref{tab:unet}}
\label{tab:summary-unet}
\centering
\footnotesize
\setlength{\tabcolsep}{2pt}
\renewcommand{\arraystretch}{1.0}
\begin{tabular}{lrrrr}
\toprule
Pair & Selection NLL\,$\downarrow$ & Completion log-loss\,$\downarrow$ & Mass $<$5\,m\,$\uparrow$ & Argmax dist.\ (m)\,$\downarrow$ \\
\midrule
U$_{\text{raw}}$ $-$ U$_{\text{raster}}$ & $-0.242$ $[-0.264,-0.213]$ & $-0.020$ $[-0.028,-0.009]$ & $+0.015$ $[+0.011,+0.020]$ & $-0.49$ $[-0.71,-0.19]$ \\
{\scriptsize\itshape matches, adj.\ $p$} & {\scriptsize sd 0.012, 8/8, $p<0.01$ $\ast$$\circ$} & {\scriptsize sd 0.008, 8/8, $p<0.01$ $\ast$$\circ$} & {\scriptsize sd 0.004, 8/8, $p<0.01$ $\ast$$\circ$} & {\scriptsize sd 0.17, 7/8, $p=0.016$ $\circ$} \\[1pt]
U$_{\text{raster}}$ $-$ C & $+0.366$ $[+0.333,+0.397]$ & $+0.018$ $[+0.005,+0.033]$ & $-0.041$ $[-0.046,-0.034]$ & $+1.76$ $[+1.58,+1.99]$ \\
{\scriptsize\itshape matches, adj.\ $p$} & {\scriptsize sd 0.014, 0/8, $p<0.01$ $\ast$$\circ$} & {\scriptsize sd 0.009, 1/8, $p=0.102$} & {\scriptsize sd 0.006, 0/8, $p<0.01$ $\ast$$\circ$} & {\scriptsize sd 0.17, 0/8, $p<0.01$ $\ast$$\circ$} \\[1pt]
U$_{\text{raw}}$ $-$ D$_{\text{raw}}$ & $+0.290$ $[+0.258,+0.325]$ & $+0.011$ $[+0.001,+0.023]$ & $-0.043$ $[-0.050,-0.037]$ & $+1.81$ $[+1.59,+2.06]$ \\
{\scriptsize\itshape matches, adj.\ $p$} & {\scriptsize sd 0.026, 0/8, $p<0.01$ $\ast$$\circ$} & {\scriptsize sd 0.004, 2/8, $p=0.484$} & {\scriptsize sd 0.006, 0/8, $p<0.01$ $\ast$$\circ$} & {\scriptsize sd 0.09, 0/8, $p<0.01$ $\ast$$\circ$} \\[1pt]
U$_{\text{raw}}$ $-$ \Gtwo{} & $+0.346$ $[+0.319,+0.379]$ & $+0.013$ $[+0.002,+0.023]$ & $-0.054$ $[-0.060,-0.049]$ & $+2.04$ $[+1.71,+2.42]$ \\
{\scriptsize\itshape matches, adj.\ $p$} & {\scriptsize sd 0.022, 0/8, $p<0.01$ $\ast$$\circ$} & {\scriptsize sd 0.008, 0/8, $p=0.164$} & {\scriptsize sd 0.004, 0/8, $p<0.01$ $\ast$$\circ$} & {\scriptsize sd 0.27, 0/8, $p<0.01$ $\ast$$\circ$} \\
\bottomrule
\end{tabular}
\end{table}

\subsection{Interface alternatives}
\label{sec:setattn}

A learned attention painter is worse than the fixed splat, and a raster-free
set decoder is worse than a raster core with no set encoder, at the shipped
recipe and at their own learning rates. The ablations of
Sec.~\ref{sec:ablation} price removals from one design and leave two questions
open that removal cannot answer: whether the bilinear splat is the right way
to write the embeddings onto the grid, and whether a grid and a convolutional
core are needed at all once the players are encoded as a set. Study 6
(Sec.~\ref{sec:prespec}) answers both with two trained arms, three seeds each
at the shipped recipe, against the locked Study 1 runs of D, \Gtwo{} and C:
four pre-declared pairs on the ten PP and PS metrics, 40 hypotheses, with
the hierarchical bootstrap over 8 matches and 3 seeds and the exact
256-pattern max-$T$ test. Table~\ref{tab:summary-setattn} summarises the family and Supplement
Table~\suppref{tab:setattn} gives every hypothesis;
the pattern in the point estimates is the result.

\paragraph{The two arms.}
D$_{\text{attn}}$ is arm~D with one operator exchanged: only the bilinear
splat is replaced by a cell-query cross-attention painter in which each of
the 7{,}072 cells queries, through fixed Fourier features of its centre,
keys built from each node's embedding and position, with the embeddings as
values, yielding the same 36 channels. The painter sees exactly what the
splat sees; the two differ in normalisation. The splat normalises per node,
so a node's contribution is conserved and most cells receive nothing
(sparse, node-conserving); the painter normalises per cell, so no cell is
empty and no node's contribution is conserved (dense, cell-normalised).
SetOnly has no raster channels and no convolutional core: the relational
encoder of \Gtwo{} encodes the 23 nodes, the 7{,}072 cell queries attend to
them through two cross-attention blocks, and a per-cell MLP gives one logit
per cell, so that the $68\times104$ logit surface feeds the same heads,
losses, calibration and evaluation as every other arm. Its widths were
chosen for the parameter count closest to \Gtwo{}'s and frozen in the
commitment, $700{,}303$ parameters, 50 below \Gtwo{}. The arm is the
set-attention family of Sec.~\ref{sec:related}~\citep{lee2019settransformer}
with the same graph inputs and relational encoder as \Gtwo{}, so that the
comparison evaluates a different decoding system, including removal of
the deterministic raster channels; it does not isolate the decoder alone. Both arms are specified in full in
Supplement Sec.~\suppref{supp:designs}.

\paragraph{A learned kernel is worse than the splat.}
Exchanging the bilinear splat for the attention painter, with encoder, core,
recipe and seeds held fixed, costs a fifth of a nat of selection NLL and a
few hundredths of local mass under both procedures; it moves the argmax farther
out and barely moves completion (Supplement Table~\suppref{tab:setattn}). The
seed-averaged selection NLL of D$_{\text{attn}}$ is $5.528$ against $5.319$
for D and $5.490$ for the raster-only arm~C: the attention-painted arm is
behind the core it feeds, and the swap costs more than discarding sub-cell
position altogether (Sec.~\ref{sec:ablation}) and more than the whole
painted package is worth (Table~\ref{tab:confirm}), with every
D$_{\text{attn}}$ head early-stopped. The painter has to learn the locality
that the splat has by construction, and at this recipe it does not: a
dense, non-conserving write of every node into every cell is worse for
selection than writing nothing.

\paragraph{The shipped model against the painter.}
\Gtwo{} keeps its margin over the painter on selection NLL, completion
log-loss and local mass under both procedures. Set against the small
increment that message passing adds on top of bilinear painting
(Sec.~\ref{sec:attribution}), about four fifths of this margin is the
painter's cost, not the relational encoder's contribution.

\paragraph{The raster-free decoder is behind the raster core.}
The raster-free arm is behind everything. Against \Gtwo{} at matched
parameter count SetOnly gives up more than half a nat of selection NLL and
some eight hundredths of completion log-loss; against the raster-only
core~C, which has no set encoder at all, it gives up a third of a nat and
nearly as much completion log-loss, all under both procedures, its
seed-averaged levels being $5.836$ nats and $0.275$ log-loss against $5.490$
and $0.209$ for C. One qualification is owed: SetOnly's selection head
reached the 50-epoch cap in all three seeds, so its selection level is what
the set decoder reaches at the shipped budget, not its ceiling, whereas its
completion deficit is one at convergence.

Two things follow for the entity-to-grid framing of Sec.~\ref{sec:intro}. The
interface is the right abstraction but the operator is not interchangeable:
the tested splat, a sparse write at continuous coordinates that conserves
each node's contribution, beats a learned attention kernel over the same
embeddings and positions by more than any ablation of the splat itself,
though the painter changes locality, sparsity and conservation at once, so
which of the three is decisive is not settled. And on this corpus the tested raster-free system is behind the grid plus
convolutional core, given the same node features and the same parameter
budget.

\begin{table}[t]
\caption{Study 6 summary: the interface alternatives at the shipped recipe: the attention-painted arm D$_{\text{attn}}$ and the raster-free SetOnly arm, three seeds each, against the locked Study 1 runs of D, \Gtwo{} and C. Intervals from the hierarchical bootstrap over 8 matches and 3 seeds ($B=10{,}000$); family of 40 hypotheses. Reading conventions: Sec.~\ref{par:reading}. Full family: Supplement Table~\suppref{tab:setattn}}
\label{tab:summary-setattn}
\centering
\scriptsize
\setlength{\tabcolsep}{2pt}
\renewcommand{\arraystretch}{1.0}
\begin{tabular}{lrrrr}
\toprule
Pair & Selection NLL\,$\downarrow$ & Completion log-loss\,$\downarrow$ & Mass $<$5\,m\,$\uparrow$ & Argmax dist.\ (m)\,$\downarrow$ \\
\midrule
D$_{\text{attn}}$ $-$ D & $+0.208$ $[+0.146,+0.271]$ & $+0.011$ $[+0.000,+0.022]$ & $-0.023$ $[-0.033,-0.013]$ & $+0.52$ $[+0.18,+0.84]$ \\
{\scriptsize\itshape matches, adj.\ $p$} & {\scriptsize sd 0.056, 0/8, $p<0.01$ $\ast$$\circ$} & {\scriptsize sd 0.011, 0/8, $p<0.01$ $\circ$} & {\scriptsize sd 0.009, 0/8, $p<0.01$ $\ast$$\circ$} & {\scriptsize sd 0.27, 0/8, $p<0.01$ $\circ$} \\[1pt]
\Gtwo{} $-$ D$_{\text{attn}}$ & $-0.260$ $[-0.317,-0.220]$ & $-0.015$ $[-0.023,-0.006]$ & $+0.029$ $[+0.025,+0.035]$ & $-0.70$ $[-0.93,-0.47]$ \\
{\scriptsize\itshape matches, adj.\ $p$} & {\scriptsize sd 0.043, 8/8, $p<0.01$ $\ast$$\circ$} & {\scriptsize sd 0.007, 8/8, $p<0.01$ $\ast$$\circ$} & {\scriptsize sd 0.002, 8/8, $p<0.01$ $\ast$$\circ$} & {\scriptsize sd 0.15, 8/8, $p<0.01$ $\ast$$\circ$} \\[1pt]
SetOnly $-$ \Gtwo{} & $+0.568$ $[+0.460,+0.681]$ & $+0.081$ $[+0.056,+0.112]$ & $-0.044$ $[-0.056,-0.035]$ & $+0.71$ $[+0.37,+1.03]$ \\
{\scriptsize\itshape matches, adj.\ $p$} & {\scriptsize sd 0.111, 0/8, $p<0.01$ $\ast$$\circ$} & {\scriptsize sd 0.027, 0/8, $p<0.01$ $\ast$$\circ$} & {\scriptsize sd 0.005, 0/8, $p<0.01$ $\ast$$\circ$} & {\scriptsize sd 0.19, 1/8, $p=0.016$ $\ast$$\circ$} \\[1pt]
SetOnly $-$ C & $+0.346$ $[+0.254,+0.435]$ & $+0.066$ $[+0.040,+0.096]$ & $-0.015$ $[-0.027,-0.005]$ & $-0.06$ $[-0.39,+0.30]$ \\
{\scriptsize\itshape matches, adj.\ $p$} & {\scriptsize sd 0.093, 0/8, $p<0.01$ $\ast$$\circ$} & {\scriptsize sd 0.027, 0/8, $p<0.01$ $\ast$$\circ$} & {\scriptsize sd 0.009, 1/8, $p=0.031$ $\ast$$\circ$} & {\scriptsize sd 0.14, 5/8, $p=1$} \\
\bottomrule
\end{tabular}
\end{table}

\paragraph{Tuned and converged.}
The qualification above is a budget, not a verdict: SetOnly's selection head
reached the cap, and both arms were trained at the shipped rates that
Supplement Sec.~\suppref{sec:lrsweep} shows favour \Gtwo{}. Study 11
(Sec.~\ref{sec:prespec}) removes both conditions for the two arms exactly as
Study 4 removed them for the controls, with the same sweep and selection
rule, seeds 2--5 at the chosen rates and the test set opened once, against
the locked Study 4 units of \Gtwo{} and C at their own selected rates: four
pre-declared pairs, 40 hypotheses, with the hierarchical bootstrap over 8
matches and 5 seeds and the exact 256-pattern max-$T$ test.
Table~\ref{tab:summary-tuned-setattn} summarises the family; Supplement
Table~\suppref{tab:tuned-setattn} gives every hypothesis. The rule chose
rates at or above the shipped ones for every head, ten times the shipped
selection rate for the raster-free arm, and no head reached the cap: what
bound the raster-free arm at the shipped recipe was the learning rate, not
the epoch cap.

\paragraph{The painter at its validation-selected learning rate.}
Tuning does not rescue the painter (Table~\ref{tab:summary-tuned-setattn}).
At its selected rate D$_{\text{attn}}$ scores $5.492$ nats of selection NLL
against $4.965$ for tuned \Gtwo{} and $5.257$ for tuned arm~C, about half a
nat behind the shipped model and a quarter behind the raster-only core under
both procedures, with less local mass and a farther argmax; completion is
level with C and slightly behind \Gtwo{}. Per-arm tuning removed roughly a third of the controls' shipped deficit
(Sec.~\ref{sec:tuned}); it improves the painter's own selection NLL by
$0.036$ nats and widens its deficit to \Gtwo{} from $0.260$ to $0.527$ nats,
so the attention-painted arm falls from level with arm~C to a quarter of a
nat behind it.

\paragraph{The raster-free arm at its validation-selected learning rate.}
Part of the raster-free arm's shipped deficit was the budget, and the rest
is not. Tuned and converged, SetOnly scores $5.476$ nats of selection NLL,
more than a third of a nat better than at the shipped recipe, yet still
about half a nat behind tuned \Gtwo{} and a fifth of a nat behind tuned
arm~C under both procedures, with completion log-loss several hundredths
worse and less local mass against both; of the third of a nat that it trailed
the raster-only core by at the shipped budget, about a third was the budget
and the rest remains at convergence. With the same node features, the same
parameter count and its own learning rate, the raster-free system is still worse than a raster and a convolutional
core with no set encoder at all, and the deficit is largest exactly where
there is no raster to fall back on, on completion.

\begin{table}[t]
\caption{Study 11 summary: the two set-attention arms of Study 6 under the Study 4 protocol, every arm at its own validation-selected learning rate, cap 200 epochs, five seeds, against the locked Study 4 runs of \Gtwo{} and C. Intervals from the hierarchical bootstrap over 8 matches and 5 seeds ($B=10{,}000$); family of 40 hypotheses. Reading conventions: Sec.~\ref{par:reading}. Full family: Supplement Table~\suppref{tab:tuned-setattn}}
\label{tab:summary-tuned-setattn}
\centering
\footnotesize
\setlength{\tabcolsep}{2pt}
\renewcommand{\arraystretch}{1.0}
\begin{tabular}{lrrrr}
\toprule
Pair & Selection NLL\,$\downarrow$ & Completion log-loss\,$\downarrow$ & Mass $<$5\,m\,$\uparrow$ & Argmax dist.\ (m)\,$\downarrow$ \\
\midrule
D$_{\text{attn}}$ $-$ \Gtwo{} & $+0.527$ $[+0.449,+0.610]$ & $+0.015$ $[+0.007,+0.024]$ & $-0.069$ $[-0.080,-0.059]$ & $+1.62$ $[+1.25,+1.97]$ \\
{\scriptsize\itshape matches, adj.\ $p$} & {\scriptsize sd 0.068, 0/8, $p<0.01$ $\ast$$\circ$} & {\scriptsize sd 0.008, 0/8, $p<0.01$ $\ast$$\circ$} & {\scriptsize sd 0.006, 0/8, $p<0.01$ $\ast$$\circ$} & {\scriptsize sd 0.30, 0/8, $p<0.01$ $\ast$$\circ$} \\[1pt]
D$_{\text{attn}}$ $-$ C & $+0.235$ $[+0.170,+0.308]$ & $+0.002$ $[-0.008,+0.013]$ & $-0.032$ $[-0.040,-0.025]$ & $+0.75$ $[+0.54,+0.94]$ \\
{\scriptsize\itshape matches, adj.\ $p$} & {\scriptsize sd 0.079, 0/8, $p<0.01$ $\ast$$\circ$} & {\scriptsize sd 0.007, 4/8, $p=1$} & {\scriptsize sd 0.008, 0/8, $p<0.01$ $\ast$$\circ$} & {\scriptsize sd 0.10, 0/8, $p<0.01$ $\ast$$\circ$} \\[1pt]
SetOnly $-$ \Gtwo{} & $+0.511$ $[+0.459,+0.566]$ & $+0.085$ $[+0.065,+0.110]$ & $-0.059$ $[-0.070,-0.049]$ & $+1.18$ $[+0.91,+1.42]$ \\
{\scriptsize\itshape matches, adj.\ $p$} & {\scriptsize sd 0.034, 0/8, $p<0.01$ $\ast$$\circ$} & {\scriptsize sd 0.022, 0/8, $p<0.01$ $\ast$$\circ$} & {\scriptsize sd 0.008, 0/8, $p<0.01$ $\ast$$\circ$} & {\scriptsize sd 0.20, 0/8, $p<0.01$ $\ast$$\circ$} \\[1pt]
SetOnly $-$ C & $+0.219$ $[+0.183,+0.261]$ & $+0.071$ $[+0.048,+0.099]$ & $-0.021$ $[-0.030,-0.014]$ & $+0.31$ $[+0.08,+0.54]$ \\
{\scriptsize\itshape matches, adj.\ $p$} & {\scriptsize sd 0.032, 0/8, $p<0.01$ $\ast$$\circ$} & {\scriptsize sd 0.025, 0/8, $p<0.01$ $\ast$$\circ$} & {\scriptsize sd 0.006, 0/8, $p<0.01$ $\ast$$\circ$} & {\scriptsize sd 0.16, 1/8, $p=0.102$} \\
\bottomrule
\end{tabular}
\end{table}

\subsection{What the surfaces rely on: probes and tracking noise}
\label{sec:robustness}

The surfaces rely on team identity, velocity and sub-cell position; under
injected tracking noise the \Gtwo{} family degrades faster than the raster
controls but keeps its selection lead. The studies above
retrain arms; the two probes here edit only the inputs of the frozen shipped
checkpoints, and carry no bootstrap interval.

\subsubsection{Inference-time probes}
\label{sec:probes}

Each probe edits the input graph or the tracking state at inference only,
reusing the shipped validation temperatures unchanged, so it measures what
the trained surfaces depend on under distribution shift.

\paragraph{Knockouts.}
Supplement Table~\suppref{tab:knockouts} gives the full knockout probe
table. Permuting the explicit pairwise edge features produces only a
$+0.0006$-nat change in selection NLL. Dropping the ball node costs $+0.101$
nats. Zeroing velocities costs $+1.29$ nats. Swapping team labels drives
completion AUC to $0.18$, below chance. Team identity and velocity dominate. Permuting the edge features moves every metric by very little, so the
shipped model draws almost nothing from the explicit pairwise edge channel
at test time; this is consistent with
the small message-passing increment above.

\paragraph{Quantisation: a sensitivity consistent with sub-cell position.}
The sharpest probe is quantisation: every player and the ball is snapped to
the centre of its $\approx$1\,m grid cell, a $105/104=1.0096$\,m by $1.0$\,m
lattice, and the snapped positions feed both the rebuilt raster and the graph
nodes, with no Gaussian noise added. It costs \Gtwo{} $+0.127$ nats
($5.2446\rightarrow5.3713$) and arm~D $+0.071$, and removes $0.0247$ and
$0.0186$ of probability mass within 5\,m. Every raster-only arm moves by at
most $0.0012$ nats (arm~C $+0.0011$, original $-0.0007$, wide $-0.0005$) and
\emph{gains} about $0.001$ of local mass. This is an inference-time
sensitivity of the trained models, not a trained causal ablation. Only the
splatting arms are materially sensitive to the lattice, as one would expect if
sub-cell precision is what the painted front end supplies; arm~D losing less
than \Gtwo{} indicates that message passing amplifies the sensitivity rather
than creating it. The quantise row is also the only selection row of the noise
table (Supplement Table~\suppref{tab:noise}) in which \Gtwo{} is not the
leading arm, and it loses that lead to arm~D, the other splatting arm, rather
than to any raster control.

\subsubsection{Tracking noise}
\label{sec:noise}

Supplement Table~\suppref{tab:noise} gives the noise curves, each level the
mean and standard deviation over $N=10$ common-random-number draws;
Sec.~\ref{sec:threats} draws their consequences. The noise model is i.i.d.\
per-frame Gaussian tracking error, $N(0,\sigma^2)$ per position component,
with an independent velocity error of amplitude $\sigma\sqrt2$, the
amplitude a one-second finite difference of two independently perturbed
frames carries. The noise is independent across frames as well as players;
a temporally correlated variant, in which the two frames the pipeline uses
carry a stationary AR(1) error pair with lag-one correlation $0.9$ or
$0.5$, is reported in Supplement Table~\suppref{tab:noise_ar1}. The same draws feed the regenerated
13-channel raster and the graph node array, so comparisons across arms are
paired within a draw. \emph{Baseline} is each arm's shipped unperturbed test
metrics and \emph{rebuild} the regeneration path at $\sigma=0$, where the
rebuilt rasters and node arrays are bit-identical to the shipped caches.

Three readings matter. First, the \Gtwo{} family is more noise-sensitive than
the raster controls: by $\sigma=1$\,m \Gtwo{} has lost $1.61$ nats of
selection NLL against the original's $1.09$. The sharper positional response
that the quantisation probe credits to the splat is also a sharper noise
response. Second, the selection advantage survives the whole range,
compressed. Under independent noise \Gtwo{}'s lead over the nearest control is
positive in every draw at every level; it compresses from $+0.52$ nats at $\sigma=0$ to
$+0.090\pm0.025$ at $1$\,m (worst draw $+0.050$) before re-widening to $+0.53$
at $4$\,m. Correlated noise does not compress it monotonically: with near-persistent error ($\rho=0.9$) the lead is $+0.200\pm0.013$ at $1$\,m but $+0.032\pm0.020$ at $2$\,m, where one of ten draws is negative against the
wide control. Against arm~C \Gtwo{} stays ahead on selection in every draw at
every level and correlation, and against arm~D in all but two of the hundred
correlated draws. Supplement Table~\suppref{tab:noise_ar1} gives the curves
against arm~C and arm~D. Third, on completion the picture is less favourable: from
$\sigma=2$\,m the ordering against the controls is within draw-to-draw noise
($-0.007\pm0.008$), and arm~D beats \Gtwo{} in 10 of 10 draws
($-0.0112\pm0.0029$).

\subsection{Development history: the retrospective three-arm comparison}
\label{sec:history}

The original comparison used one run per arm at a shared recipe on the
eight-match holdout. Its selection-NLL advantage over the wide control was
$0.5156$ nats; later per-arm learning-rate selection reduced that margin to
about a third of a nat. These records are retained as development history,
not as independent confirmation, and the evaluation events behind them are
in Sec.~\ref{sec:records} and Sec.~\ref{sec:threats}. Supplement Sec.~\suppref{supp:history} gives the full results,
multi-seed control extension, learning curves, and historical figures;
Sec.~\ref{sec:threats} explains the implications. The primary evidence is
the cross-validation, external retraining, and tuned comparisons above.

\subsection{Value surfaces}
\label{sec:value}

Value is where the paper's null models bite: \Gtwo{} beats the constant-zero
predictor on squared error, and no arm beats it on absolute error. In this
corpus PE is empirically the hardest of the three tasks: $92.6\%$ of test
rewards are exactly zero, so value is a downstream integration result rather
than a headline. It rests on only $459$ test passes with nonzero reward, $62$
of them negative. Supplement Table~\suppref{tab:value} gives the EPV
comparison (Eq.~\eqref{eq:epv}) over all 6{,}196 test rows against two null
models. The \emph{constant-zero} predictor, the null model for a target that
is zero on most rows, attains MSE $0.003482$ and MAE $0.010666$ for every arm.
The \emph{frozen} baseline predicts the two outcome-conditioned
training-branch means, gated by each arm's own calibrated PP surface; skill
scores are relative to it. The \emph{graph-gate} column re-gates every arm
with \Gtwo{}'s calibrated PP surface, holding the gate common and letting only
the value branches differ; the \emph{oracle-gate} column substitutes the
realised pass outcome for the predicted probability.

Value performance is metric-dependent. Among the learned systems \Gtwo{}
leads both controls on EPV error, and against frozen per-branch means four of
five arms carry positive MSE skill, up to $+0.102$; \Gtwo{} also beats the
constant-zero predictor on MSE. But no arm beats constant zero on MAE ($0.01067$ against the best model's
$0.01605$), every arm being $1.5$--$2.0\times$ worse. The original control's
value heads show no detectable skill over the frozen constants (MSE difference
$+1.44\times10^{-5}$, $[-7.10\times10^{-5},+7.71\times10^{-5}]$, bootstrap
$p=0.659$). The gate is nearly irrelevant. Re-gating with \Gtwo{}'s PP closes
only $6.2\%$ of the \Gtwo{}-versus-original MSE gap and $9.7\%$ of the
\Gtwo{}-versus-wide gap, while an oracle gate buys \Gtwo{} just $2.6\%$ of its
own MSE; the value branches rather than the completion gate do what work there
is. On the 459 nonzero-reward rows every arm carries positive skill over the
frozen baseline on both MSE and MAE. The positive tail is resolved (\Gtwo{}'s
precision@25 of $0.64$ is a $10\times$ lift over base rate); the negative tail
is resolved by no arm, the best precision@25 being $0.08$ against only 62
negative rows. The wide control's advantages on imbalance-aware sign metrics
carry intervals excluding zero (Supplement Table~\suppref{tab:tail},
Sec.~\ref{sec:threats}). Sign accuracy is not
evidence for a \Gtwo{} advantage.

\subsection{Compute cost}
\label{sec:compute}

Compute cost depends on the representation and on the batching regime.
Supplement Table~\suppref{tab:compute} retains the original measurements,
and a post-hoc audit measures all six principal arms, D$_{\mathrm{raw}}$
included, under one implementation, GPU, precision and timing protocol
(Supplement Table~\suppref{tab:compute-audit}), separating single-example
latency from batch-32 amortised cost and excluding preprocessing and data
transfer. At batch size one, arm~C takes $2.800$\,ms per forward pass,
D$_{\mathrm{raw}}$ $4.127$\,ms and \Gtwo{} $6.559$\,ms; amortised over a
batch of 32 the three cost $0.087$, $0.128$ and $0.201$\,ms per pass. Raw
painting is therefore cheaper than \Gtwo{} but not free relative to the
raster-only core. \Gtwo{}'s smaller counted FLOP total than the wide control
does not translate into lower measured latency, because FLOP counting omits
some scatter and indexing work. These measurements price a cost--accuracy
trade-off; the recommendation of Sec.~\ref{sec:intro} follows from them
together with the accuracy results.

\subsection{The surfaces themselves}
\label{sec:surfaces}

\begin{figure}[tbp]
\centering
\includegraphics[width=0.78\textwidth]{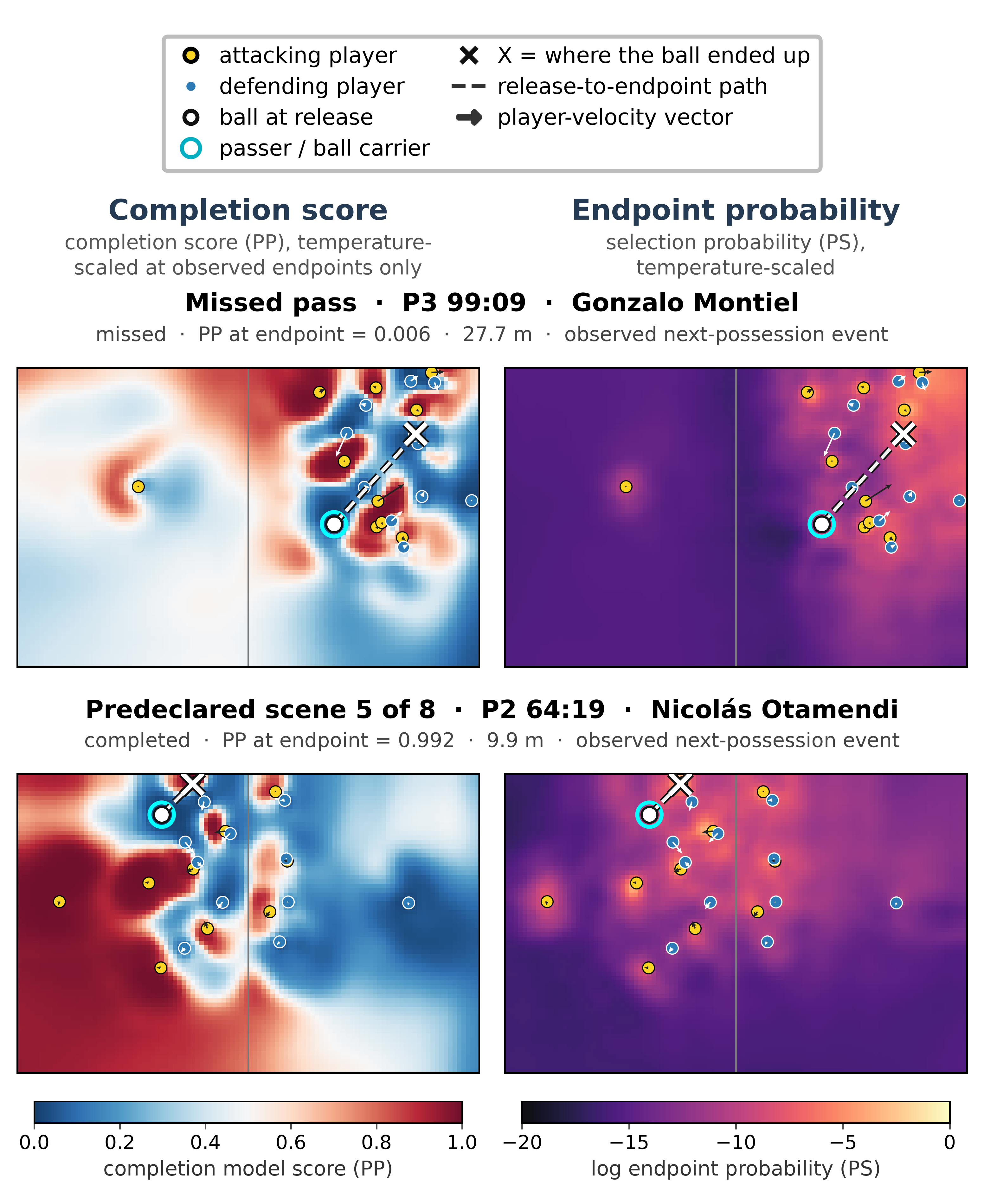}
\caption{Two scenes of the 2022 World Cup final, one row per scene. Row 1 is
\emph{outcome-selected}: of the final's missed passes of at least $10$\,m,
the one with the lowest completion score at its endpoint --- a $27.7$\,m
extra-time attempt whose endpoint is the global minimum of the completion
map ($0.006$). Row 2 is \emph{pre-declared} in the sealed evaluation record:
a completed second-half pass ($0.992$). Left, the model's completion score
over candidate destinations, temperature-scaled at observed endpoints and an
extrapolation elsewhere. Right, the temperature-scaled distribution over the
ball's next destination, on a log scale clipped at $e^{-20}$; it sums to one
over the 7{,}072 cells. Yellow: attackers; blue: defenders; cyan ring: the
player on the ball; X: where the ball ended up}
\label{fig:gallery}
\end{figure}

Fig.~\ref{fig:gallery} shows what the model draws for two real scenes of the
World Cup final, pairing a failure with a success. The completion score
collapses inside the opponent's block and behind pressing lines; the
selection surface concentrates on nearby, unpressured destinations. Both
panels are temperature-scaled model outputs: the selection surface is a
distribution the loss shapes everywhere, whereas the completion surface is
calibrated only at observed endpoints and is an extrapolation elsewhere
(Sec.~\ref{sec:estimands}). The eight pre-declared scenes of the shipped evaluation were chosen by a
purely chronological rule (equal index strata of the final's eligible-pass
sequence, midpoint of each stratum). They were hashed into the test commitment
record before any test forward pass and rendered only after the metric lock;
the second displayed scene comes from that pre-declared set, and the first,
as the caption states, was chosen after the scores were known. Neither scene
is quantitative validation.

\section{Limitations and Threats to Validity}
\label{sec:threats}

Each threat is stated here once; the record-level detail behind each
statement is given in Supplement Sec.~\suppref{supp:limitations}.

\textbf{The shipped three-arm comparison is retrospective.} The arms were
evaluated sequentially and the controls were not externally timestamped
before the first \Gtwo{} evaluation, so the records cannot establish that
their design choices were independent of that earlier result. We call it
retrospective and controlled, not pre-specified: the label is ours, the
adaptive-reuse danger \citet{dwork2015holdout}'s. It also rests on one
training run per arm, which Study 2 (Supplement Sec.~\suppref{sec:multiseed})
addresses. The cross-validation, the tuned comparison and the external
replication now carry the paper's claim, and only the external replication
scores data that took no part in development; what depends on the original
split throughout is the choice of architectures, recipe and grid, which no
study here evaluates design-blind.

\textbf{The \Gtwo{} arm opened the test set twice.} Its log
(Sec.~\ref{sec:records}) records two commitment/evaluation cycles; the first
was followed by an automated integrity failure that quarantined the
transaction and triggered retraining, and the shipped evaluation is the
second. The log cannot establish whether any first-cycle output was
inspected; each control shows one commitment and zero recovery events. The
complete first-cycle block is printed in full (Supplement
Sec.~\suppref{app:firstcycle}): the discarded opening was the better draw on
completion and on combined EPV, and the shipped cycle leads only on calibrated selection NLL, by a few
thousandths of a nat against control gaps of half a nat or more, and
marginally on PE-succ MSE, so selection toward a favourable draw is not a
plausible reading of these records.

\textbf{The attribution ladder was first run post hoc.} Its original
single-seed form ran after the test set had been opened and was therefore
exploratory. Sec.~\ref{sec:attribution} reports the pre-specified
replacement; two of the single-seed findings did not replicate and have been
withdrawn (Supplement Sec.~\suppref{app:ladder}).

\textbf{The replications reuse the holdout, and therefore do not cure
adaptivity.} The pre-specified multi-seed studies of
Secs.~\ref{sec:attribution}--\ref{sec:setattn} and~\ref{sec:tuned}, and
Study 2, run on the previously used eight-match split. What they cure is
real: analysis, metric family, comparison pairs and multiplicity correction
were fixed in a signed commitment before training and the checkpoints sealed
before a single opening. What they cannot cure is that those matches had
already been opened and that the hypotheses they test were derived from that
earlier look; no reusable-holdout mechanism of the kind
\citet{dwork2015holdout} develop was used. Read them as evidence that the
attribution is stable across initialisations and was not selected after the
fact, not as independent confirmation on unseen data. The cross-validation
and the external replication reproduce the headline on matches never before
scored as test data, but the recipe and the architectures of both were
chosen on the original split, and the external corpus is seven matches from
a single second provider. The ladder prices \emph{removals} from one trained
design at one budget rather than mapping an architecture space: arm~D shows
what happens when message passing is deleted here, not that no relational
encoder could do better.

\textbf{What the intervals and tests cover.} Intervals condition on the
fitted models, the dataset construction and the retained design choices.
Match resampling accounts for within-match clustering but does not model
dependence from recurring teams or from overlapping cross-validation
training sets, and one seed per cross-validation fold does not estimate
training-seed uncertainty, so the reported uncertainty does not cover every
source of variation under deployment or a new tournament. The sign-flip
test's error control rests on sign-exchangeability of the paired match-level
deltas under the joint null, which match grouping motivates but does not
guarantee, and strong familywise control needs the corresponding subset-null
conditions as well. Multiplicity is controlled within each study's family,
not across the adaptive sequence of thirteen studies; prior holdout exposure
and later reporting choices preclude reading the sequence as prospective
confirmation. The full families are retained to make that sequence
inspectable.

\textbf{Roughly a third of the matched-budget selection margin disappears
under per-arm learning-rate tuning.} With every arm tuned and converged over
five seeds, the selection margin against the capacity-matched control falls
from about half a nat to a third (Sec.~\ref{sec:tuned}); the direction is
unchanged. The tuning grid is coarse, and which control is the stronger
selector is itself learning-rate dependent.

\textbf{Calibration is unresolved.} No calibration-error comparison in any
study survives correction. The metric is badly conditioned here, about two
thirds of every arm's calibrated completion predictions falling in the
single top bin (Supplement Sec.~\suppref{app:calibration}), and calibration
is in any case established only where the labels are: outcomes at unchosen
cells are never observed, so neither whole-grid calibration nor the
counterfactual validity of the surface away from the selected destination
has been established. The external study adds a second failure mode: frozen
across a provider boundary, \Gtwo{}'s calibrated selection likelihood falls
behind the raster-only core in all seven matches and behind the raw-feature
painter, on the untempered score as well, while its localisation advantage
survives. The raw-feature painter's better transfer is consistent with a
learned encoder fitted to the source distribution without proving it, and
the \Gtwo{}$-$D$_{\text{raw}}$ contrast changes the per-node encoder and
message passing together, so it isolates neither.

\textbf{Value performance depends on the metric.} No arm beats a
constant-zero predictor on EPV MAE, the original control's value heads show
no detectable skill over two frozen constants, no arm resolves the negative
reward tail, and the wide control exceeds \Gtwo{} on imbalance-aware sign
metrics (Sec.~\ref{sec:value}).

\textbf{Measurement error sets a floor, and \Gtwo{} is more sensitive to it.}
Broadcast-derived positions carry provider-dependent error reaching several
metres~\citep{crang2026validity,mills2026events}. Under the i.i.d.\ Gaussian
perturbation of Sec.~\ref{sec:noise} the \Gtwo{} family degrades faster than
the raster controls: its selection lead stays positive in every draw at
every level of independent noise, and in all but one draw under correlated
noise, but has compressed to a small fraction of its noise-free size by
one metre of per-coordinate error, and on completion the ordering against
the controls is within draw-to-draw noise from two metres, with arm~D ahead
of \Gtwo{} in every draw. Here $\sigma$ is additional error per coordinate,
whereas provider validations often report two-dimensional error against
ground truth. The perturbations are added to measurements that already
contain unknown error, and the targets are held fixed. They cannot be
converted directly into an expected advantage for a provider at a reported
RMSE. The experiments measure sensitivity to the specified extra noise, not
recovery of true player positions.

\textbf{What the secondary comparisons do not settle.} Each ablation and
raster control of Secs.~\ref{sec:ablation} and~\ref{sec:raster} is one
change to one design at one budget, three seeds, on the reused holdout, and
each confounds something: nearest-cell painting changes the interpolation as
well as the placement, the offset raster carries a per-type within-cell mean
rather than each node's own position, and the finer raster halves every
convolution's receptive field. They establish deficits for the raster
constructions tested, not for every raster model. Failure to reject the
raw-versus-MLP contrast in the three-seed study does not establish
equivalence; the combined learned-content increment is resolved in the larger
primary comparisons, where message passing alone was not isolated. The
attention painter and the raster-free decoder are specific implementations
rather than exhaustive representatives of their families, and each changes
several properties at once, so neither comparison isolates a single inductive
bias; a kernel constrained to be local and mass-conserving was not tried.
The external raw-feature study reuses the external matches of Study 7 and
strengthens attribution on that corpus rather than adding an independent
replication.

\textbf{What the present design leaves open.} The canonical U-Net of
Study 13 is a second core-invariance check at the shared recipe, and the
published surface core of
\citet{overmeer2025revisiting}, with its risk/reward split, was not
reproduced. A second tournament from the same provider, which would separate
provider shift from league shift, is not in this paper; nor are a
threshold-sensitivity analysis and a blinded manual audit of the endpoint
pipeline.

\textbf{Target semantics and validation.} The selection target is an
observed next-event endpoint, not the passer's intent; for incomplete passes
the intended target is
unobserved~\citep{robberechts2023unxpass,anzer2022expected}. The value
target is model-generated, an XGBoost~\citep{chen2016xgboost} expected-goals
estimator fitted on StatsBomb open data~\citep{statsbomb_opendata}, so the
labels carry estimation error of the kind \citet{vanarem2026quality}
quantify. No blinded human audit establishes endpoint accuracy for either
corpus: the per-gate audit verifies the extraction accounting, not the
semantic correctness of each target, and the automated corroboration of the
separate benchmark project is neither new human ground truth nor a
validation result of this paper. Indicator-level validation, expert-paired
benchmarks such as~\citet{overmeer2025revisiting}, is absent.
Provider-smoothed input positions have no established causal filtering
guarantee here, so the study is retrospective prediction from supplied
snapshots, not a validated live system. The external replication is seven matches from one other provider, with the
value heads unscored and an estimated event--frame alignment; it establishes
that the ordering and the size of the interface effect transfer when the
arms are retrained, and that frozen likelihood ordering does not. Its
endpoints are $91.2\%$ receiver captures against $0.21\%$ on the World Cup,
and the post-hoc stratification of Sec.~\ref{sec:external} shows the
external margin concentrated in that stratum, so on this corpus the
interface effect is not separated from a label rule that places endpoints at
tracked player positions; the World Cup corpus, whose labels are almost
entirely next-event, is what shows the effect does not require that rule. Independent release-time and endpoint
annotation, construction-threshold sensitivity, and downstream decision
evaluation remain necessary for stronger practical claims.

\section{Conclusion}
\label{sec:conclusion}

Writing each player's features onto the pitch at the player's continuous
position improves endpoint-supervised surfaces on every core and corpus we
trained it on. Relative to the same raster-only core, the full front end reduces
selection NLL by $0.251$ nats across the World Cup folds, $0.292$ nats under
per-arm learning-rate selection, and $0.239$ nats after external retraining,
ahead in every match of each comparison. The same write pays on the original
SoccerMap and on a canonical U-Net. The operator is standard bilinear
scatter-add; the finding is that the interface, not what is written through
it, carries most of the gain.

Painting the nine raw node features with no learned encoder retains
approximately three quarters of that gain. The learned encoder and message
passing add a further $0.065$--$0.067$ nats in the cross-validation and tuned
comparisons, an increment that is real but small. For a new deployment we
would therefore start from raw painting, which has no front-end parameters
and costs $4.127$\,ms per example against \Gtwo{}'s $6.559$\,ms at batch size
one (Supplement Table~\suppref{tab:compute-audit}), and add the learned
front end where in-distribution selection likelihood is the objective and
the provider is fixed. Frozen across a provider boundary, the ordering
reverses: the raster-only core has the best selection likelihood of the
three, raw painting is numerically ahead of \Gtwo{} without an adjusted
max-$T$ rejection, and only retraining restores the interface effect.

The evidence concerns reconstructed observed endpoints. Reused holdouts,
overlapping folds, provider-dependent tracking error, and unobserved outcomes
at alternative destinations bound what the intervals cover. The value task
remains secondary and metric-dependent. Independent endpoint validation and
evaluation of downstream decisions are needed before these surfaces can be
treated as calibrated assessments of hypothetical passes.

\section*{Statements and Declarations}
\addcontentsline{toc}{section}{Statements and Declarations}

\paragraph{Funding.}
No funding was received for conducting this study. Computing resources were
the authors' own.

\paragraph{Competing interests.}
The authors have no competing interests to declare that are relevant to the
content of this article. The World Cup data were obtained under the
provider's research licence; neither author has a financial relationship
with PFF FC, StatsBomb or the IDSSE authors.

\paragraph{Data availability.}
The raw PFF FC World Cup tracking and event data~\citep{pff2024worldcup}
are obtained from the provider on request under its terms of use and are not
redistributed by us; reconstructing that corpus requires a copy obtained
directly from the provider. The public IDSSE source
data~\citep{bassek2025idsse} are available under CC~BY~4.0 from their
authors, and the expected-goals estimator uses public StatsBomb
data~\citep{statsbomb_opendata}; data provided by StatsBomb. The per-row locked predictions,
study commitment and lock records, and analysis tables that underlie every
reported number are retained by the authors and are available from the
corresponding author on reasonable request.

\paragraph{Code availability.}
The source code is not publicly available. The supplement specifies every
arm, study design, admission gate and scoring rule in enough detail to
re-implement them.

\paragraph{Author contributions.}
Conceptualization: K.G.\ and O.G. Methodology and model development: K.G.
Software, experimental design, formal analysis, investigation and
visualization: O.G. Original draft: O.G. Review, editing and final
manuscript preparation: K.G. Both authors reviewed and approved the
manuscript.

\paragraph{Use of AI tools.}
Claude (Anthropic) assisted with drafting and refactoring analysis and
figure scripts, deriving multi-seed orchestration code from the training
engine, and restructuring and editing manuscript text. Codex (OpenAI)
assisted with consistency checks, figure generation, source-to-result
checks, and a post-hoc implementation and compute audit that left all locked
predictive results unchanged. These uses extend beyond copy-editing, and the
authors reviewed and verified the output of both tools. The tools are not
authors; the human authors are responsible for the study design, the
verification and interpretation of results, and approval of the final
manuscript.

\paragraph{Prior dissemination.}
No part of this manuscript has been published, presented, posted or
submitted elsewhere. A companion manuscript by the same authors describing
the public pass-surface corpus of Sec.~\ref{sec:external} as a benchmark is
in preparation; it shares data and extraction code with this paper but no
claim, result or text (Sec.~\ref{sec:intro}).

\paragraph{Ethics approval and consent.}
Not applicable: the study uses licensed and public sports data on
professional matches and involves no recruitment, intervention, or
collection of new participant data.

\begingroup
\footnotesize
\bibliographystyle{apalike}
\bibliography{refs}
\endgroup

\end{document}